\documentclass[accepted]{uai2021}

\usepackage[american]{babel}
\usepackage{natbib}
\usepackage{algorithm}
\usepackage{algorithmic}
\usepackage{amsthm}
\usepackage{booktabs}
\usepackage{caption}
\usepackage{graphicx}
\usepackage{mathtools}
\usepackage{microtype}
\usepackage{multirow}
\usepackage{subcaption}
\usepackage{tikz}
\usepackage{xr-hyper}

\usepackage{amsmath,amsfonts,bm}

\def\figref#1{figure~\ref{#1}}

\def\eqref#1{equation~\ref{#1}}

\def\1{\bm{1}}

\def\vb{{\bm{b}}}

\def\vw{{\bm{w}}}
\def\vx{{\bm{x}}}
\def\vy{{\bm{y}}}
\def\vz{{\bm{z}}}

\def\mH{{\bm{H}}}
\def\mI{{\bm{I}}}
\def\mJ{{\bm{J}}}

\def\mX{{\bm{X}}}

\DeclareMathAlphabet{\mathsfit}{\encodingdefault}{\sfdefault}{m}{sl}
\SetMathAlphabet{\mathsfit}{bold}{\encodingdefault}{\sfdefault}{bx}{n}
\newcommand{\tens}[1]{\bm{\mathsfit{#1}}}

\def\tH{{\tens{H}}}

\def\gA{{\mathcal{A}}}

\def\gD{{\mathcal{D}}}

\def\gH{{\mathcal{H}}}
\def\gI{{\mathcal{I}}}

\def\gN{{\mathcal{N}}}

\def\sR{{\mathbb{R}}}

\newcommand{\E}{\mathbb{E}}

\newcommand{\Var}{\mathrm{Var}}

\newcommand{\Cov}{\mathrm{Cov}}

\DeclareMathOperator*{\argmax}{arg\,max}
\DeclareMathOperator*{\argmin}{arg\,min}

\newtheorem*{definition}{Definition}

\graphicspath{ {./figures/} }
\newcommand\defeq{\stackrel{\textrm{def}}{=}}

\title{Measuring Data Leakage in Machine-Learning Models\\
with Fisher Information}

\author[1]{\href{mailto:Awni Hannun <awni@fb.com>?Subject=Your UAI 2021 paper}{Awni~Hannun}{}}
\author[1]{Chuan~Guo}
\author[1]{Laurens~van~der~Maaten}
\affil[1]{
    Facebook AI Research
}

\begin{document}
\maketitle

\begin{abstract}
Machine-learning models contain information about the data they were trained
on. This information leaks either through the model itself or through
predictions made by the model. Consequently, when the training data contains
sensitive attributes, assessing the amount of information leakage is
paramount. We propose a method to quantify this leakage using the Fisher
information of the model about the data. Unlike the worst-case \emph{a
priori} guarantees of differential privacy, \emph{Fisher information loss}
measures leakage with respect to specific examples, attributes, or
sub-populations within the dataset. We motivate Fisher information loss
through the Cram\'{e}r-Rao bound and delineate the implied threat model.
We provide efficient methods to compute Fisher information loss for output-perturbed
generalized linear models. Finally, we empirically validate Fisher
information loss as a useful measure of information leakage.
\end{abstract}

\section{Introduction}
\label{sec:intro}

Machine-learning models trained on sensitive data are often made public. Even
when the models are not explicitly released, they may be implicitly leaked from
their predictions~\citep{papernot2017practical, tramer2016stealing}.
Undeniably, these models contain information about the data they were trained
on. Without mitigating measures, training set membership can often be
inferred~\citep{shokri2017membership, yeom2018privacy}, and, in some cases,
sensitive attributes or even whole examples can be
extracted~\citep{carlini2019secret, carlini2020extracting,
fredrikson2014privacy, fredrikson2015model}.

Assessing the information leaked from models about their training data is
commonly done with techniques such as differential
privacy~\citep{dwork2006calibrating}. Differential privacy successfully avoids
the ``just a few'' failure mode of more heuristic privacy assessments, in which
privacy is protected for many but not all individuals~\citep{dwork2014}.
This comes at the cost of a worst-case assessment, leading to large
differences in the vulnerability of individuals to privacy attacks. Also, a
mismatch exists between the protection of differential privacy, which is
relative to participation in a dataset, and privacy attacks, which can take
advantage of the absolute information leaked from a trained
model~\citep{carlini2019secret, long2018understanding}. Furthermore,
differential privacy implicitly degrades when correlations exist in the
dataset~\citep{ghosh2016inferential, humphries2020differentially,
kasiviswanathan2008thesemantics, liu2016dependence}.

We propose an example-specific and correlation-aware measure of
data leakage using Fisher information. The quantity, which we term
\emph{Fisher information loss}, can assess the leakage of a model
about various subsets of the full dataset. This includes, for example, assessments at the
granularity of individual attributes, individual examples, groups of examples,
or the full training set. We show, via the Cram\'{e}r-Rao bound, that under
specific assumptions the ability of an adversary to estimate the underlying data
from a model with bounded Fisher information loss is limited. We also demonstrate
that, unlike differential privacy, Fisher information loss does not
implicitly degrade when data is correlated.

We derive tractable computations of Fisher information loss in the case of
Gaussian noise perturbation with generalized linear models.
We empirically validate Fisher information loss with experiments on four datasets.
We further demonstrate that Fisher information loss accurately
captures examples susceptible to attribute inversion attacks.

The ability of Fisher information loss to measure per-example privacy loss means
it can be used as the basis for algorithms that aim to achieve fairness in
privacy~\citep{cummings2019compatibility, ekstrand2018privacy}. We
demonstrate this by developing an algorithm that balances Fisher information
loss for individuals in the training set, thereby resolving the problem that
subgroups may have ``disparate vulnerability'' to privacy attacks
~\citep{yaghini2019disparate}.

\section{Related Work}
\label{sec:related_work}

This work builds on and complements a significant body of prior work in
assessing the privacy of a model with respect to the data it was trained on.
This includes approaches which obfuscate the original data such as
$k$-anonymity~\citep{samarati1998} and
$l$-diversity~\citep{machanavajjhala2007diversity}, information theoretic
criteria~\citep{agrawal2001design}, and, perhaps the most commonly studied,
differential privacy~\citep{dwork2006calibrating} and its more recent
variations~\citep{dong2019gaussian, mironov2017renyi}. While these
methods often provide rigorous privacy guarantees, most of them do not
explicitly bound the \emph{inferential power} of an adversary, and correlations
in the training dataset can be exploited by an adversary to mount a successful
inference attack~\citep{machanavajjhala2007diversity,li2007t,ghosh2016inferential,
liu2016dependence}. Moreover, these definitions do not offer simple and flexible
approaches to measure the information leaked by a model about varying subsets
of the training data. This can lead to unequal vulnerability to privacy attacks
despite attempts to protect privacy~\citep{yaghini2019disparate}.

Because these privacy assessment techniques do not identify vulnerability at
the level of the individual or sub-population, prior work exists to quantify
the susceptibility of such subgroups to privacy attacks.
\citet{farokhi2020modelling} propose an information theoretic measure of the
vulnerability of individual examples to membership inference attacks. Less
rigorous heuristics have also been studied~\citep{long2017towards,
long2018understanding}.  \citet{carlini2019secret} propose a heuristic which
can infer the susceptibility of data to model inversion attacks.

This work also builds upon prior studies of Fisher information as a measure of
privacy. An early study analyzed loss of Fisher information in a general
randomized response framework~\citep{anderson1977}. More recently, the
relationship of privacy with Fisher information to estimator error through the
use of the Cram\'{e}r-Rao bound was investigated~\citep{farokhi2017fisher}. In
particular, \citet{farokhi2017fisher} turn to Fisher information as a practical
alternative to differential privacy in protecting data collected from smart
power meters. We build from these works in several directions, including a
broader application to generalized linear models as well as the use of Fisher
information loss in an algorithm which can provide fairness in data leakage.

\section{Fisher Information Loss}
\label{sec:fisher_information_loss}

Let $\gD = \{(\vx_1, y_1), \ldots, (\vx_n, y_n)\}$ be a training dataset of $n$
examples with $\vx_i \in \sR^d$ and scalar target $y_i$. We denote by $\gA(\gD)$ a
randomized learning algorithm which outputs a hypothesis $h$ from a predefined
hypothesis space $\gH$. Treating the hypothesis as a random variable,
$p_\gA(h \mid \gD)$ is the probability
density of $h$ given $\gD$ for the randomized algorithm $\gA(\gD)$. We
denote by $\gI_h(\gD) \in \sR^{n(d+1) \times n(d+1)}$ the Fisher information
matrix (FIM) defined by
\begin{equation}
 \gI_h(\gD) = -\E_h \left[ \nabla_\gD^2 \log p_\gA(h \mid \gD) \right]
\end{equation}
where $\nabla^2_\gD$ yields the matrix of second derivatives of $\log p_\gA(h
\mid \gD)$ with respect to the values in $\gD$, and the expectation is taken
over the randomness in $\gA(\gD)$. The FIM, $\gI_h(\gD)$, is a measure of the
information that the hypothesis $h$ contains about the training data $\gD$.
Hence, we use the FIM to measure the information loss from releasing the output
$h$ of a single evaluation of $\gA(\gD)$.

\begin{definition}[Fisher information loss]
\label{def:fil}
  We say that $h \sim \gA(\gD)$ has Fisher information loss (FIL) of $\eta$
  with respect to $\gD$ if
  \begin{equation}
    \label{eq:fil}
    \left\| \gI_h(\gD) \right\|_2 \le \eta^2,
  \end{equation}
  where $\|\gI_h(\gD)\|_2$ denotes the $2$-norm, or largest singular value, of
  the FIM. A smaller $\eta$ means $h$ contains less Fisher information about
  the training data, $\gD$.
\end{definition}

{\bf Motivation.} Fisher information is a classical tool used in
statistics for lower bounding the variance of an
estimator~\citep{lehmann2006theory}. We utilize this property to demonstrate
that a small FIL implies a large variance for any unbiased estimate of the
data.

Let $\vz = [\vx_1^\top, y_1, \ldots, \vx^\top_n, y_n] \in
\sR^{n (d+1)}$ be the vector formed by concatenating the examples in $\gD$, and
$z \in \vz$ an arbitrary element. If the FIL of $h$ with respect to $\gD$ is
bounded by $\eta$, then for any unbiased estimator $\hat{z}$ of $z$, we have:
\begin{equation}
  \Var(\hat{z}) \ge \frac{1}{\eta^2}.
  \label{eq:var_lower}
\end{equation}
Hence, a smaller FIL implies a larger variance in any unbiased attempt to infer
$z$. Equation \ref{eq:var_lower} follows directly from the
Cram\'{e}r-Rao bound. Indeed, under a fairly relaxed regularity
condition~\citep{kay1993fundamentals}, for any unbiased estimator $\hat{\vz}$ of
the data $\vz$, the Cram\'{e}r-Rao bound states that:
\begin{equation}
  \E\left[(\hat{\vz} - \vz)(\hat{\vz} - \vz)^\top\right] \succeq \gI_h(\vz)^{-1},
\end{equation}
where $A \succeq B$ if the matrix $A-B$ is positive semidefinite. Since the
estimator is unbiased, this implies the covariance of $\hat{\vz}$ is similarly
bounded:
\begin{equation}
  \label{eq:cov_fim_bound}
  \Cov(\hat{\vz}) \succeq \gI_h(\vz)^{-1}.
\end{equation}
If $\gA(\gD)$ has an FIL of $\eta$ then \eqref{eq:fil} implies
$\gI_h(\vz)^{-1}_{ii} \ge 1 / \eta^2$ for all $i$, and from
\eqref{eq:cov_fim_bound}, $\Var(\hat{z}) \ge 1 / \eta^2$ follows.

{\bf Correlated data.} Fisher information loss also provides some security in
the presence of intra-dataset correlations. If a model has an FIL bounded by
$\eta$ with respect to the training data (in vector form $\vz$), then the
covariance matrix for any unbiased estimator $\hat{\vz}$ of that data is
bounded by $\|\Cov(\hat{\vz})\|_2 \ge 1 / \eta^2$. This limits the ability of
an adversary using an unbiased estimator to infer relative differences between
elements in the dataset.

\begin{figure}
  \centering
  \includegraphics[width=\columnwidth]{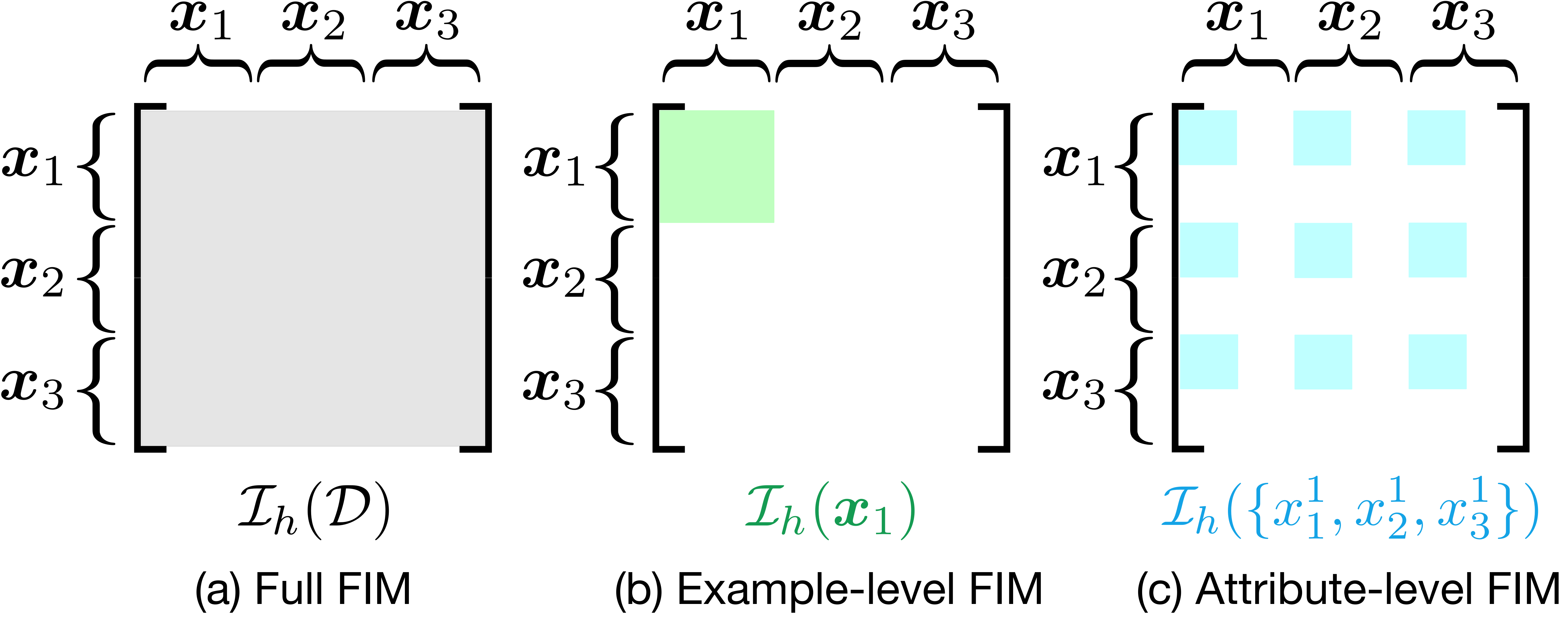}
  \caption{The FIM for different subsets of the training data.}
  \label{fig:fim_figure}
  \vspace{-1ex}
\end{figure}

\subsection{Properties of FIL}
\label{sec:properties_fil}

{\bf Subsets.} In many cases we are interested in measuring FIL with
respect to a single example $(\vx_i, y_i) \in \gD$.  In this case, we can
compute the example-specific FIM by selecting the corresponding
entries of the full FIM. Given the FIM for the vector represented data $\vz$,
computing the FIM for the $i$-th example amounts to selecting the submatrix of
size $(d+1) \times (d+1)$ with upper left corner $\gI_h(\vz)_{i(d+1), i(d+1)}$.
Similarly, computing the FIM for a specific attribute over all examples,
$\{x_i^j\}_{i=1}^n$, amounts to constructing the submatrix by selecting the
corresponding entries from the full FIM. See figure \ref{fig:fim_figure} for an
illustration. In general, we can compute the FIM for any subset of elements of
$\gD$ by selecting their corresponding entries from the full FIM.

{\bf Composition.} The Fisher information, and hence FIL, compose
additively. Given $k$ independent evaluations of $\gA(\gD)$ each with an FIL of
$\eta$, the combined FIL is at most $\sqrt{k}\eta$. More generally, given an
evaluation of $k$ unique but independent randomized algorithms $\{h_i \sim
\gA_i(\gD) \mid i=1, \ldots, k\}$ each with an FIM of $\gI_{h_i}(\gD)$, the FIM
with respect to all of the $h_i$'s is given by $\gI_{h_1, \ldots, h_k}(\gD) =
\sum_{i=1}^k \gI_{h_i}(\gD)$~\citep{lehmann2006theory}. Hence, if
$\|\gI_{h_i}(\gD)\|_2 \le \eta_i^2$, then the combined FIL for all $h_i$ about
$\gD$ is at most $(\sum_{i=1}^k \eta_i^2)^{1/2}$.  This follows from the triangle
inequality applied to the matrix $2$-norm.

{\bf Closed under post-processing.} As in differential privacy, FIL is
closed under post-processing. If $\|\gI_h(\gD)\|_2 \le \eta^2$, then for any
function of the hypothesis, $g(h)$, $\|\gI_{g(h)}(\gD)\|_2 \le \eta^2$.  This
follows since the matrix $\gI_h(\gD) - \gI_{g(h)}(\gD)$ is positive
semidefinite for any function $g(h)$~\citep{schervish2012theory}.

{\bf Threat model.} The bounds on an adversary's ability to estimate the
data apply under several assumptions about the problem setting and the
adversary:
\begin{itemize}
  \item The required regularity condition is satisfied so that the FIM exists
    and the Cram{\'e}r-Rao bound applies~\citep{kay1993fundamentals}. Namely, the
    expected value of the score function of the density $p_\gA(h \mid \gD)$
    should be $0$ for all $\gD$:
    \begin{equation}
      \E_h \left[ \frac{\partial \log p_\gA(h \mid \gD)}{\partial \gD} \right] = 0,
    \end{equation}
    which essentially states that the derivative and integral commute:
    \begin{align*}
      \E_h \left[ \frac{\partial \log p_\gA(h \mid \gD)}{\partial \gD} \right] &=
      \int_h \frac{p_\gA(h \mid \gD)}{p_\gA(h \mid \gD)} \frac{\partial p_\gA(h \mid \gD)}{\partial \gD} d h \\
      &= \frac{\partial}{\partial \gD} \int_h p_\gA(h \mid \gD) d h = 0.
    \end{align*}
    This condition is satisfied for the models and privacy mechanisms we
    consider.
  \item We assume the adversary is limited to unbiased estimators of the
    unknown data. While achieving lower variance estimators is possible, this
    means the estimator will have to incur bias. Denoting by $\psi(\vz) =
    \E[\hat{\vz}]$ the expected value of the estimator as a function of the
    unknown data, the Cram\'{e}r-Rao bound for biased estimators generalizes
    to:
    \begin{equation}
      \Cov(\hat{\vz}) \succeq \mJ_\psi \gI_h(\vz)^{-1} \mJ_\psi^\top,
    \end{equation}
    where $\mJ_\psi$ is the Jacobian of $\psi$ with respect to $\vz$. This
    gives a bound on the variance:
    \begin{equation}
      \label{eq:crb_variance}
      \Var(\hat{z}_i) \ge \mJ_{\psi, i}^\top \gI_h(\vz)^{-1} \mJ_{\psi, i},
    \end{equation}
    where $\mJ_{\psi, i}$ is the $i$-th row of
    $\mJ_\psi$~\citep{lehmann2006theory}. Because mean squared error decomposes
    into a variance and a bias term, we observe:
    \begin{equation}
      \label{eq:bias_variance_bound}
      \E\left[\|\hat{\vz} - \vz\|^2_2 \right] \ge \mJ_{\psi, i}^\top \gI_h(\vz)^{-1} \mJ_{\psi, i} + \|\E\left[ \hat{\vz}\right] - \vz\|^2_2.
    \end{equation}
    When estimating individual elements $z_i \in \vz$,
    \eqref{eq:bias_variance_bound} reduces to:
    \begin{equation}
      \label{eq:bias_variance_bound_scalar}
      \E\left[(\hat{z}_i - z_i)^2 \right] \ge \left(\frac{\partial \psi}{\partial z_i}\right)^2\gI_h(\vz)^{-1}_{ii} + (\E\left[ \hat{z}_i\right] - z_i)^2.
    \end{equation}
    We observe from \eqref{eq:crb_variance} that there are two ways to reduce
    the variance of the estimator: 1) increase the Fisher information or 2)
    reduce the sensitivity of the estimator to the value being estimated. If
    the Fisher information is constant, then the reduction in variance must
    come from the second option, which incurs bias. In the scalar case, per
    \eqref{eq:bias_variance_bound_scalar}, any attempt to reduce the variance
    of the estimator will result in higher bias and will not yield a smaller
    mean squared error. However, in the multivariate case  bounded FIL alone
    does not guarantee a bounded mean squared error. James-Stein type shrinkage
    can result in estimators with a smaller increase in bias than the
    corresponding decrease in variance relative to an unbiased
    estimator~\citep{lehmann2006theory}.

  \item When estimating FIL for subsets of $\gD$, we assume the remainder of
    the data is \emph{known} by the adversary. This assumption arises from the
    fact that selecting elements of the full FIM corresponding to the subset is
    equivalent to computing the FIM for that subset directly. Using the FIM of
    the subset to compute FIL makes no assumptions about the remainder of
    $\gD$. In fact, if the remainder of the data is assumed to be unknown or
    partially known, FIL is still a valid upper bound on information leakage.
    Intuitively, the adversary \emph{gains at least as much information} about
    the target subset with knowledge of the remaining data. We derive a
    mathematically precise statement of this claim below.

    Suppose the adversary aims to infer example $(\vx_1, y_1)$ without knowledge of
    the remaining data. Let $\vz$ be the full data vector and let $g(\vz)$ be a
    function that selects the first $d+1$ dimensions of $\vz$ corresponding to
    $(\vx_1, y_1)$. For an unbiased estimator $(\hat{\vx}_1, \hat{y}_1)$ of
    $(\vx_1, y_1)$, the Cram\'{e}r-Rao bound under the parameter transformation $g$
    is given by:
    \begin{equation}
    \Cov((\hat{\vx}_1, \hat{y}_1)) \succeq \mJ_g^\top \gI_h(\vz)^{-1} \mJ_g,
    \end{equation}
    where $\mJ_g$ is the Jacobian of $g$. Simplifying this expression gives
    $\Cov((\hat{\vx}_1, \hat{y}_1)) \succeq [\gI_h(\vz)^{-1}]_{0:d,0:d}$.

    To finish the derivation, one can use the matrix block-inversion
    formula~\citep{petersen07thematrix} to obtain:
    \begin{equation}
    [\gI_h(\vz)^{-1}]_{0:d,0:d} \succeq [\gI_h(\vz)_{0:d,0:d}]^{-1},
    \end{equation}
    the latter of which is the FIM for $(\vx_1, y_1)$ under the assumption that the
    remaining data is known. Thus $\Cov((\hat{\vx}_1, \hat{y}_1)) \succeq \eta_1^2$
    where $\eta_1$ is the FIL for $(\vx_1, y_1)$. This derivation can be easily
    generalized to any subset of the training data.
\end{itemize}

\section{Computing FIL}
\label{sec:computing_fil}

{\bf Learning setting.} We assume a linear model with parameters $\vw
\in \sR^d$ and minimize the regularized empirical risk:
\begin{equation}
  \label{eq:erm}
  \vw^* = f(\gD) \defeq \argmin_\vw \sum_{i=1}^n \ell(\vw^\top \vx_i, y_i) + \frac{n\lambda}{2} \| \vw \|_2^2.
\end{equation}
Furthermore, we assume that the loss $\ell(\vw^\top \vx_i, y_i)$ is convex and
twice differentiable. We denote by $f(\gD)$ the minimizer, $\vw^*$, of
\eqref{eq:erm} as a function of the dataset $\gD$. When computing FIL at the
example level, we let $f_i(\vx, y)$ be the minimizer of \eqref{eq:erm} as a
function of the $i$-th data point:
\begin{equation}
  \label{eq:erm_example}
  f_i(\vx, y) \defeq \argmin_\vw \sum_{j \ne i} \ell(\vw^\top \vx_j, y_j) + \ell(\vw^\top \vx, y) + \frac{n\lambda}{2} \| \vw \|_2^2.
\end{equation}

{\bf Output perturbation.} The definition of FIL in equation \ref{def:fil} only
applies to a randomized learning algorithm $\gA$ with a differentiable density
function. To obtain such a randomized learning algorithm from the minimizer $\vw^*$
in equation \ref{eq:erm}, we adopt the Gaussian mechanism from differential
privacy. A randomized algorithm $\gA$ satisfies $(\epsilon,
\delta)$-differential privacy with respect to dataset $\gD$ if:
\begin{equation}
p(\gA(\gD) = h) \le e^\epsilon p(\gA(\gD') = h) + \delta
\end{equation}
for all $\gD$ and $\gD'$ which differ by one example and all hypothesis $h \in
\gH$~\citep{dwork2006calibrating}. The Gaussian mechanism adds zero-mean
isotropic Gaussian noise to the parameters $\vw^*$. For a given $\epsilon$ and
$\delta$, the standard deviation $\sigma$ can be chosen such that the Gaussian
mechanism satisfies $(\epsilon, \delta)$-differential
privacy~\citep{dwork2014}.

{\bf Fisher information loss.} Let $\gA(\gD) = f(\gD) + \vb$ be the
output-perturbed learning algorithm, where $\vb \sim \gN(0, \sigma^2 \mI)$.
The FIM of $\vw' \sim \gA(\gD)$ is given by:
\begin{equation}
  \label{eq:fim_gaussian_mech}
  \gI_{\vw'}(\gD) = \frac{1}{\sigma^2} \mJ_f^\top \mJ_f,
\end{equation}
where $\mJ_f \in \sR^{d \times n(d + 1)}$ is the Jacobian of $f(\gD)$ with
respect to $\gD$. The Jacobian $\mJ_f$ captures the sensitivity of the
minimizer $\vw^*$ with respect to the training dataset $\gD$. See
Appendix~\ref{apx:fim_gaussian_derivation} for a simplified derivation of
\eqref{eq:fim_gaussian_mech} and~\citet{kay1993fundamentals} for a more
rigorous and general treatment. The FIL of the Gaussian mechanism with scale
$\sigma$ is then:
\begin{equation}
  \label{eq:fil_gaussian_mech}
  \eta = \frac{1}{\sigma} \|\mJ_{f}\|_2.
\end{equation}

{\bf Jacobian of the minimizer.} We aim to compute the Fisher information loss
of the Gaussian mechanism in \eqref{eq:fil_gaussian_mech} at the example level.
This requires the Jacobian $\mJ_{f_i}$ of $f_i(\vx, y)$,
\eqref{eq:erm_example}, with respect to $(\vx, y)$ evaluated at $(\vx_i,
y_i)$. For a convex, twice-differentiable loss function $\ell(\vw^\top \vx, y)$,
the Jacobian $\mJ_{f_i}$ evaluated at $(\vx_i, y_i)$ is given by:
\begin{equation}
  \label{eq:minimizer_jacobian}
  \mJ_{f_i} \bigg\rvert_{\vx_i, y_i} = -\mH_{\vw^*}^{-1} \nabla_{\vx, y} \nabla_\vw \ell(\vw^{*\top} \vx_i, y_i).
\end{equation}
The Hessian is computed over the full dataset $\mH_{\vw^*} = \sum_{i=1}^n
\nabla^2_{\vw} \ell(\vw^{*\top}\vx_i, y_i) + n \lambda \mI$. The term $\nabla_{\vx, y}
\nabla_\vw \ell = [\nabla_\vx \nabla_\vw \ell, \nabla_y \nabla_\vw \ell] \in
\sR^{d \times (d+1)}$ is the Jacobian of $\nabla_\vw \ell$ with respect to
$(\vx, y)$ with entries given by:
\begin{equation}
  \left(\nabla_\vx \nabla_\vw \ell\right)_{ij} = \frac{\partial (\nabla_\vw \ell)_i}{\partial \vx_j}
    \;\;\textrm{and} \;\;
    \left( \nabla_y \nabla_\vw \ell \right)_i = \frac{\partial (\nabla_\vw \ell)_i}{\partial y}.
\end{equation}
A derivation is given in Appendix~\ref{apx:jacobian_general}.

\subsection{Model-Specific Derivations}
\label{sec:jacobians}

{\bf Linear regression.} In linear regression, the loss function is:
\begin{equation}
  \ell(\vw^\top \vx, y) = \frac{1}{2} \left(\vw^\top \vx - y\right)^2.
\end{equation}
Let $\mX = [\vx_1, \ldots, \vx_n]^\top$ be the design matrix and $\vy = [y_1,
\ldots, y_n]^\top$ be the vector of labels. The gradient of $\ell$ with respect
to $\vw$ is:
\begin{equation}
\nabla_\vw \ell(\vw^\top \vx, y) = (\vw^\top \vx - y) \vx,
\label{eq:linear_gradient}
\end{equation}
Taking Jacobian of equation \ref{eq:linear_gradient} with respect to $(\vx, y)$
gives:
\begin{equation}
\nabla_\vx \nabla_\vw \ell = \vx \vw^{\top} + \left(\vw^\top \vx - y\right) \mI,
\text{  and  }
\nabla_y \nabla_\vw \ell = -\vx.
\end{equation}
Using $\mH_{\vw^*} = \mX^\top \mX + n \lambda \mI$, we can then combine the
above with equation \ref{eq:minimizer_jacobian} to obtain
the partial Jacobians $\mJ_{f_i, \vx}$ and $\mJ_{f_i, y}$. Finally,
using \eqref{eq:fil_gaussian_mech},
the FIL $\eta_i$ for example $(\vx_i, y_i)$ with the Gaussian mechanism is
$\eta_i~=~\frac{1}{\sigma} \|\mJ_{f_i}\|_2$ where
$\mJ_{f_i} = \left[\mJ_{f_i, \vx}, \; \mJ_{f_i, y}\right]$.

{\bf Logistic regression.} The binary logistic regression loss is:
\begin{equation}
  \ell(\vw^\top \vx, y) = -y \log s(\vw^\top \vx) - (1-y) \log (1 - s(\vw^\top \vx))
\end{equation}
where $y \in \{0, 1\}$ and $s(a) = 1 / (1 + \exp(-a))$. The gradient of the
loss with respect to $\vw$ is:
\begin{equation}
  \label{eq:logistic_gradient}
  \nabla_\vw \ell(\vw^\top \vx, y) = (s(\vw^\top \vx) - y) \vx,
\end{equation}
and the Hessian is:
\begin{equation}
  \label{eq:logistic_hessian}
  \nabla^2_\vw \ell(\vw^\top \vx, y) = s(\vw^\top \vx)(1 - s(\vw^\top \vx)) \vx \vx^\top.
\end{equation}
The partial Jacobians of the gradient in \eqref{eq:logistic_gradient} with
respect to $(\vx, y)$ are:
\begin{align}
  \nabla_\vx \nabla_\vw \ell &= s(\vw^\top \vx) (1 - s(\vw^\top \vx)) \vx \vw^\top + (s(\vw^\top \vx) - y)\mI,
  \intertext{and}
  \nabla_y \nabla_\vw \ell &= -\vx.
\end{align}
We can combine the partial Jacobians with the Hessian in
\eqref{eq:logistic_hessian} to form the full Jacobian $\mJ_{f_i}$ in
\eqref{eq:minimizer_jacobian}.

{\bf Computational considerations.} Computing the FIM for a given example
$(\vx_i, y_i)$ requires computing the Jacobian $\mJ_{f_i}$ in
\eqref{eq:minimizer_jacobian}.  For both linear and logistic regression,
computing the inverse Hessian, $\mH_{\vw^*}^{-1}$ requires $O(d^2n)$
operations, assuming $n \ge d$. Since the Hessian is the same for all examples
in $\gD$, the cost may be amortized over $n$ evaluations of per-example FIL.
Constructing the per-example FIM requires $O(d^3)$ operations, and the largest
singular value can computed efficiently by the power method. In the more
general case, with $\vz \in \sR^m$ computing $\gI_{\vw'}(\vz)$ requires
$O(dm^2)$ operations, which is quadratic in the size of $\vz$. If $\vz$ is the
full dataset, for example, then $m = (d+1) n$ which results in $O(d^3 n^2)$
operations to compute the FIM.

\begin{algorithm}[t]
\caption{Iteratively reweighted Fisher information loss.}
\label{alg:irfil}
\begin{algorithmic}[1]
\STATE \textbf{Input}: Data set $\gD$, loss function $\ell(\cdot)$, number of
  iterations $T$, and noise scale $\sigma$.
\STATE Initialize sample weights $\omega_i^0 \gets 1$.
\FOR {$t \gets 1$ to $T$}
  \STATE $\vw^* \gets \argmin_\vw \sum_{i=1}^n \omega_i^{t-1} \ell(\vw^\top \vx_i, y_i) + \frac{n \lambda}{2} \|\vw\|^2_2$.
  \STATE $\vw' \gets \vw^* + \vb$ \; where \; $\vb \sim \gN(0, \sigma^2 \mI)$.
  \STATE $\eta_i \gets \left(\|\gI_{\vw'}(\vx_i, y_i)\|_2\right)^{1/2}$.
  \STATE $\omega_i^t \gets \frac{n\omega_i^{t-1} / \eta_i}{\sum_{i=1}^n \omega_i^{t-1} / \eta_i}$. \label{alg:line:update}
\ENDFOR
\STATE \textbf{Return}: The private weights $\vw'$.
\end{algorithmic}
\end{algorithm}

\begin{figure*}[t]
  \centering
  \begin{subfigure}{0.24\textwidth}
  \centering
  \includegraphics[width=1.0\textwidth]{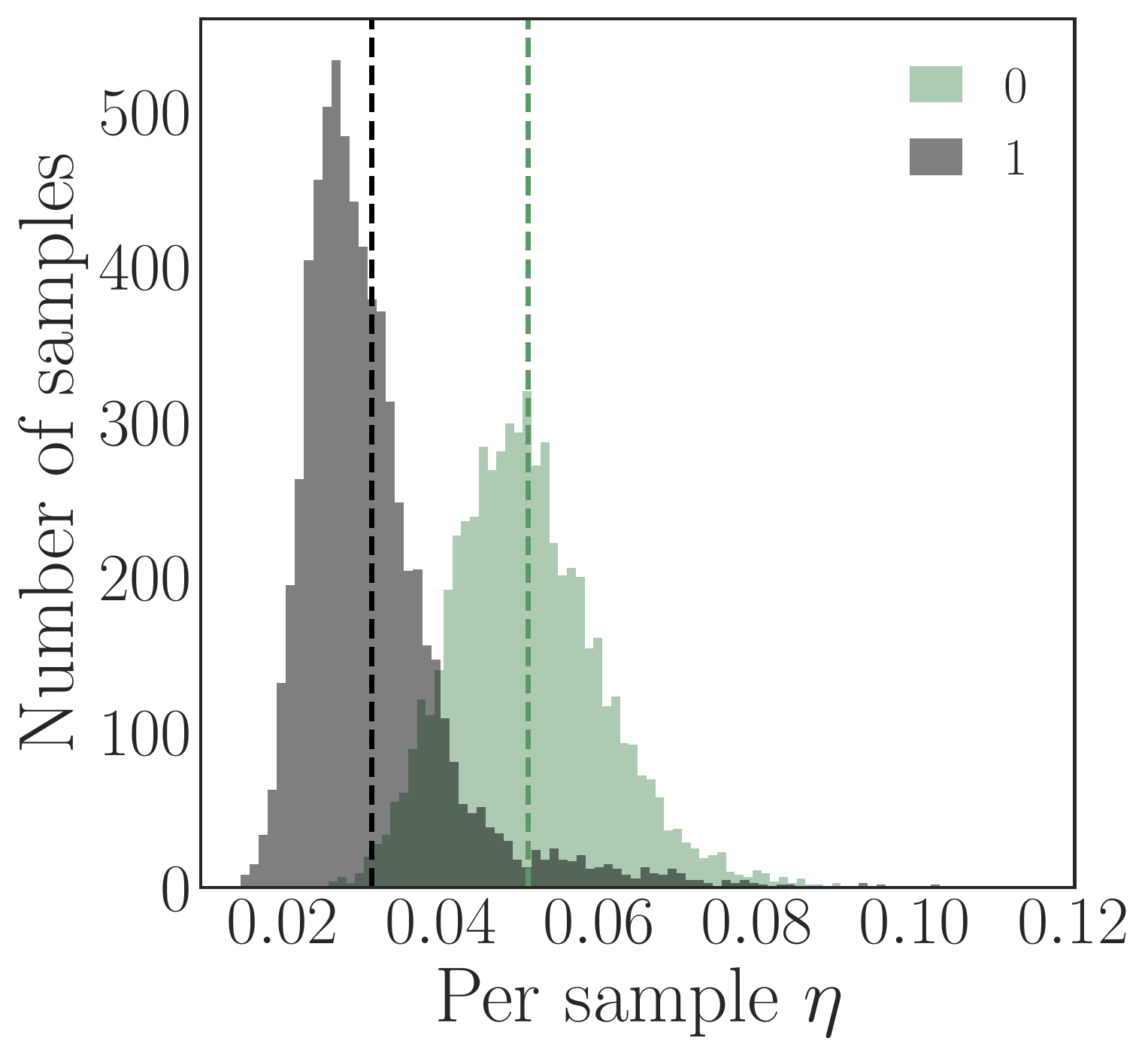}
  \caption{Linear, MNIST}
  \end{subfigure}
  \begin{subfigure}{0.24\textwidth}
  \centering
  \includegraphics[width=1.0\textwidth]{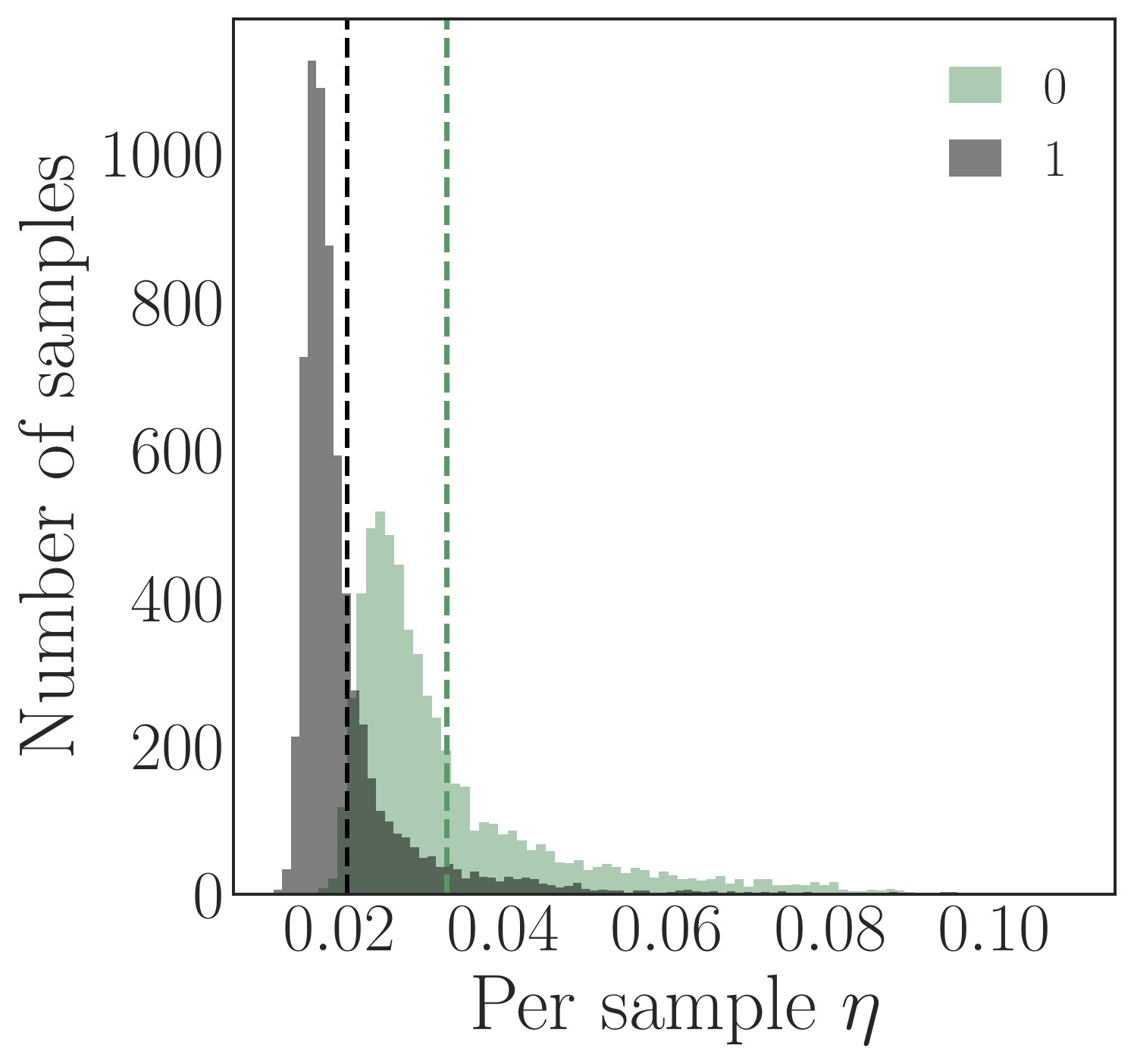}
  \caption{Logistic, MNIST}
  \end{subfigure}
  \begin{subfigure}{0.24\textwidth}
  \centering
  \includegraphics[width=1.0\textwidth]{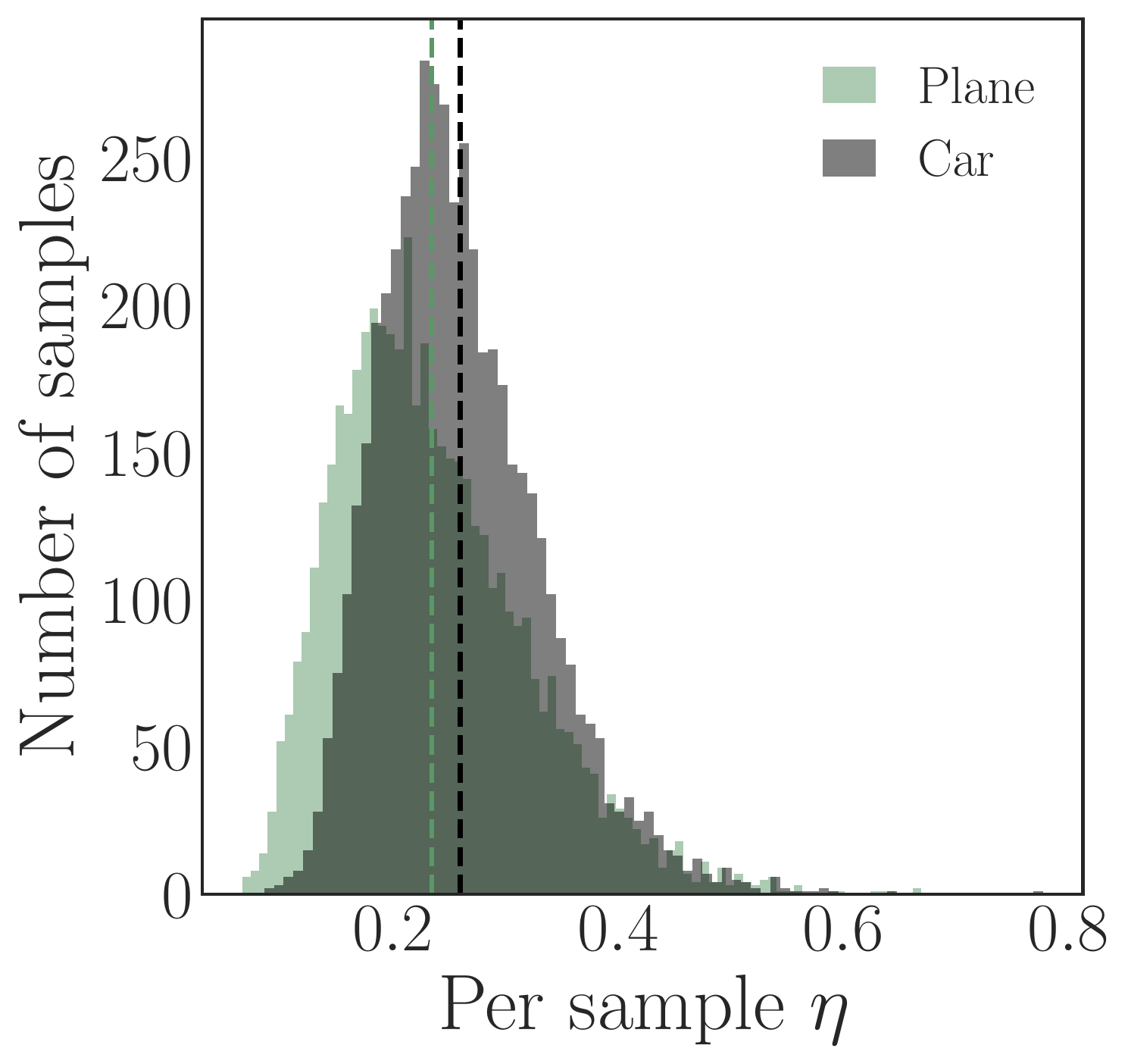}
  \caption{Linear, CIFAR-10}
  \end{subfigure}
  \begin{subfigure}{0.24\textwidth}
  \centering
  \includegraphics[width=1.0\textwidth]{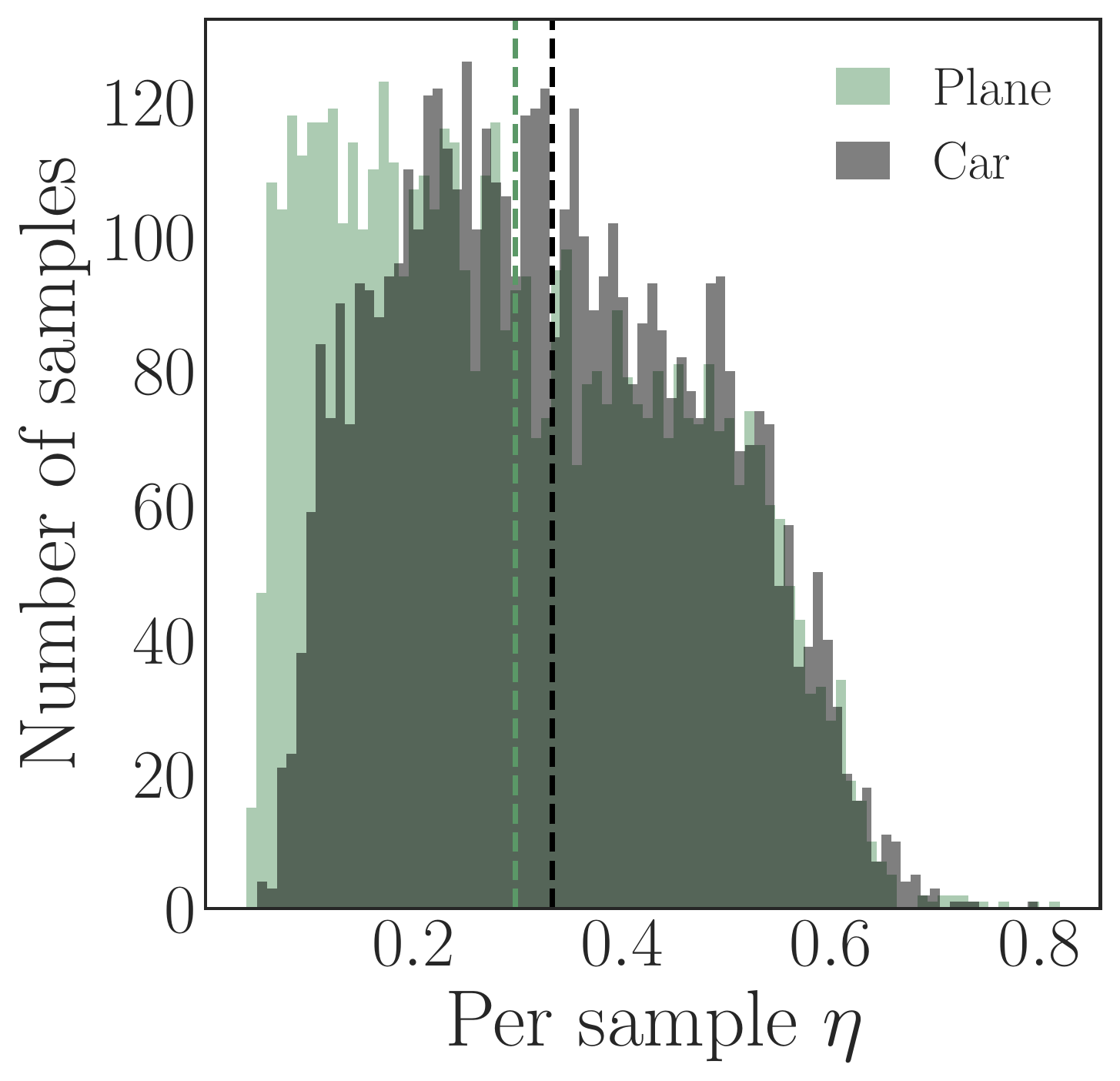}
  \caption{Logistic, CIFAR-10}
  \end{subfigure}
  \caption{Histograms of per-example $\eta$ separated by class label for the
  MNIST and CIFAR-10 training sets for linear and logistic regression. Each
  class label's mean $\eta$ is denoted by the dashed vertical line.}
  \label{fig:fil_histograms}
\end{figure*}

\section{Iteratively Reweighted FIL}
\label{sec:reweighted_fil}

Prior work has shown that existing privacy mechanisms fail to provide equitable
protection against privacy attacks for different
subgroups~\citep{yaghini2019disparate}. We address this issue with iteratively
reweighted Fisher information loss (IRFIL; Algorithm~\ref{alg:irfil}), which
yields a model with an equal per-example FIL across all examples in $\gD$. This
is done by re-weighting the per-example surrogate loss $\ell(\vw^\top \vx_i,
y_i)$ over repeated computations of the minimizer $\vw^*$.

After the first iteration in Algorithm~\ref{alg:irfil}, the weight for the
$i$-th example is inversely proportional to the initial $\eta_i$. At successive
iterations, the per-example weight $\omega_i^{t}$ is multiplicatively updated
by a value inversely proportional to the current model's FIL. The normalization
on line~\ref{alg:line:update} is primarily for numerical stability, keeping the
weights from shrinking too rapidly. Without $L_2$ regularization, the resulting
$\eta_i$ are invariant to the norm of the weights. With $L_2$ regularization,
the normalization helps to keep the ratio of the primary objective to the
regularization term constant.

If $\eta_1 = \ldots = \eta_n = \eta$ are constant, then
Algorithm~\ref{alg:irfil} has converged since the update on
line~\ref{alg:line:update} yields a fixed point:
\begin{equation}
    \omega_i^t = \frac{n \omega_i^{t-1} / \eta}{\sum_{i=1}^n \omega_i^{t-1} / \eta} =
    \frac{n\omega_i^{t-1}}{\sum_{i=1}^n \omega_i^{t-1}} = \frac{n \omega_i^{t-1}}{n} = \omega_i^{t-1},
\end{equation}
where the second-to-last equality follows since the $\omega_i^{t-1}$ were
normalized to sum to $n$ at the previous iteration.

Including the weights in computing $\gI_{\vw'}(\vx_i, y_i)$ via
\eqref{eq:minimizer_jacobian} is a straightforward application of the chain
rule:
\vspace{-2ex}
\begin{equation}
  \mJ_{f_i} = -\omega_i \left(\textrm{diag}(\omega_1, \ldots, \omega_n) \mH_{\vw^*}\right)^{-1} \nabla_{\vx, y} \nabla_\vw \ell.
\end{equation}
\vspace{-3ex}

\begin{figure*}[t]
  \begin{subfigure}{0.5\textwidth}
  \centering
  \includegraphics[width=1.0\textwidth]{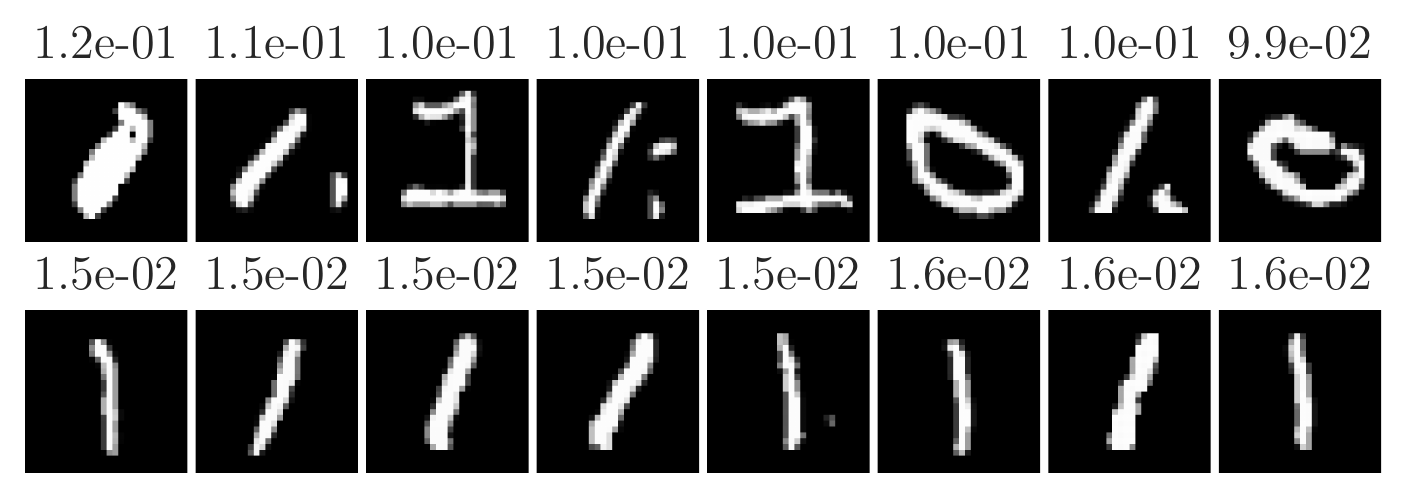}
  \caption{Linear, MNIST}
  \end{subfigure}
  \begin{subfigure}{0.5\textwidth}
  \centering
  \includegraphics[width=1.0\textwidth]{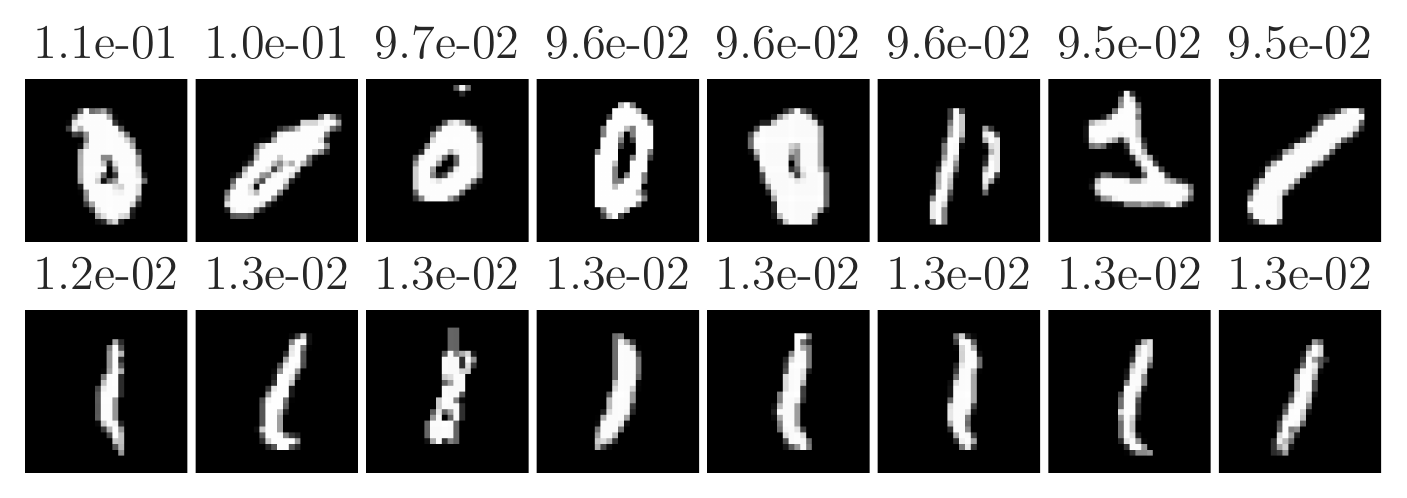}
  \caption{Logistic, MNIST}
  \end{subfigure}
  \par\medskip
  \begin{subfigure}{0.5\textwidth}
  \centering
  \includegraphics[width=1.0\textwidth]{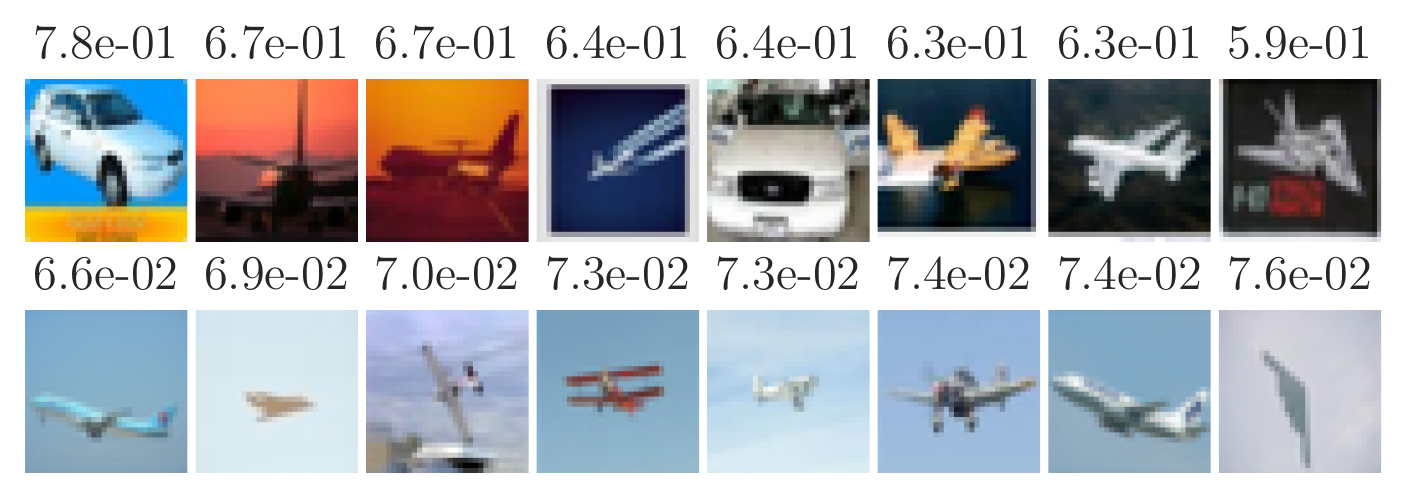}
  \caption{Linear, CIFAR-10}
  \end{subfigure}
  \begin{subfigure}{0.5\textwidth}
  \centering
  \includegraphics[width=1.0\textwidth]{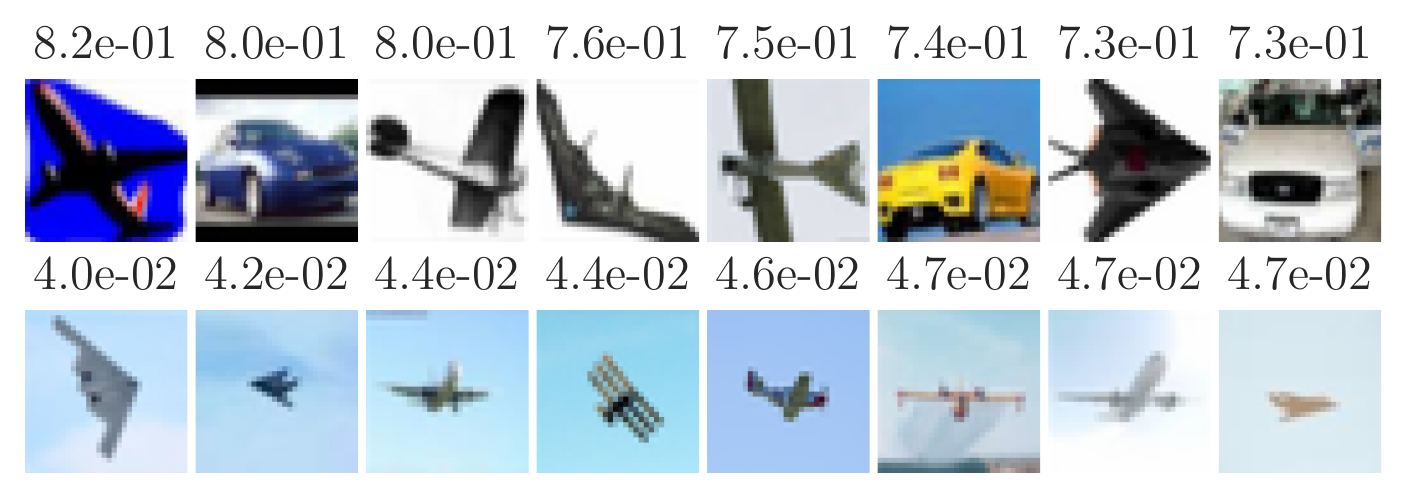}
  \caption{Logistic, CIFAR-10}
  \end{subfigure}
  \caption{The eight images with the smallest and largest $\eta$ over the MNIST
  and CIFAR-10 training sets for linear and logistic regression. The number
  above each individual image is the corresponding $\eta$.}
  \label{fig:fil_images}
  \vspace{-1ex}
\end{figure*}

\section{Experiments}
\label{sec:experiments}

We perform experiments with linear and logistic regression models on
MNIST~\citep{lecun1998} and CIFAR-10~\citep{krizhevsky2009}. For attribute
inversion attacks, we use the ``IWPC" dataset from the Pharmacogenetics and
Pharmacogenomics Knowledge Base~\citep{international2009estimation} and the
``Adult" dataset from the UC Irvine data repository~\citep{dua2019}. Code to
reproduce our results is available at
\url{https://github.com/facebookresearch/fisher_information_loss}.

\subsection{Experimental Setup}
{\bf Datasets.} On MNIST we perform binary classification of the digits $0$ and
$1$ using a training dataset of 12,665 examples. On CIFAR-10, we classify
images of planes from cars using a training dataset with 10,000 examples. We
normalize all inputs to lie in the unit ball, $\max_i \|\vx_i\|_2 \le 1$, and
then project each input using PCA onto the top twenty principal components for
the corresponding dataset.

The IWPC dataset requires estimating the stable dose of warfarin (an oral
anticoagulant) based on clinical and genetic traits. We use the same
preprocessed dataset as~\citet{yeom2018privacy}. Of the 4,819 examples, we
randomly split 20\% into a test set and use the remaining 80\% for the training
set. The UCI Adult dataset requires classifying individuals with income above
or below \$50,000. We remove examples with missing features, leaving 30,162
examples in the training set and 15,060 in the test set. For the purpose of
attribute inference, the marital status features are combined into a single
category of married or unmarried and relationship features are
removed~\citep{mehnaz2020black}. For both IWPC and UCI Adult, nominal features
are converted to one-hot vectors with the last value dropped to avoid perfect
collinearity in the encoding. Numerical features are centered to zero mean and
unit variance. We do not otherwise preprocess the datasets, yielding a total of
14 features per example for IWPC and 86 features per example for UCI Adult.

\label{sec:experimental_setup}

{\bf Hyperparameters.} Unless otherwise stated, for linear regression we do not
apply $L_2$ regularization. For logistic regression, $\lambda$ is set to the
largest value such that the training set accuracy is the same up to two
significant digits as that of linear regression. For linear regression, the
targets are $y_i \in \{-1, 1\}$, and we compute the exact minimizer. For
logistic regression, we use limited-memory BFGS to compute the minimizer. We
typically report $\eta$ assuming a Gaussian noise scale of $\sigma = 1$; hence
$\eta = \|\mJ_{f}\|_2$.

\subsection{Validating Fisher Information Loss}

Histograms of the per-example $\eta$ for both MNIST and CIFAR-10 are shown in
\figref{fig:fil_histograms}, separated by output class label. On MNIST the
histograms have distinct modes. For both linear and logistic regression, the
mode at the larger value is for the digit $0$, implying that the model in
general contains more information about images of $0$ than of $1$. While the
modes are not as distinct on CIFAR-10, the class-specific means are still
separate, with class label ``car" having a larger $\eta$ in general.

Figure~\ref{fig:fil_images} shows the eight images with the largest and
smallest $\eta$ in each of the four settings. The images with the smallest
$\eta$ are consistent with the class means in table~\ref{tab:eta_label_mean}.
For MNIST, these correspond to the digit $1$ written in a very typical
manner. For CIFAR-10, the smallest $\eta$ images are small planes on a blue-sky
background. As expected, the images with the largest $\eta$ are much more
idiosyncratic. Of the 100 largest $\eta$ examples for logistic
and linear regression, 24 overlap for MNIST and 34 overlap for CIFAR-10.

\begin{figure}[t]
  \centering
  \includegraphics[width=1.0\linewidth]{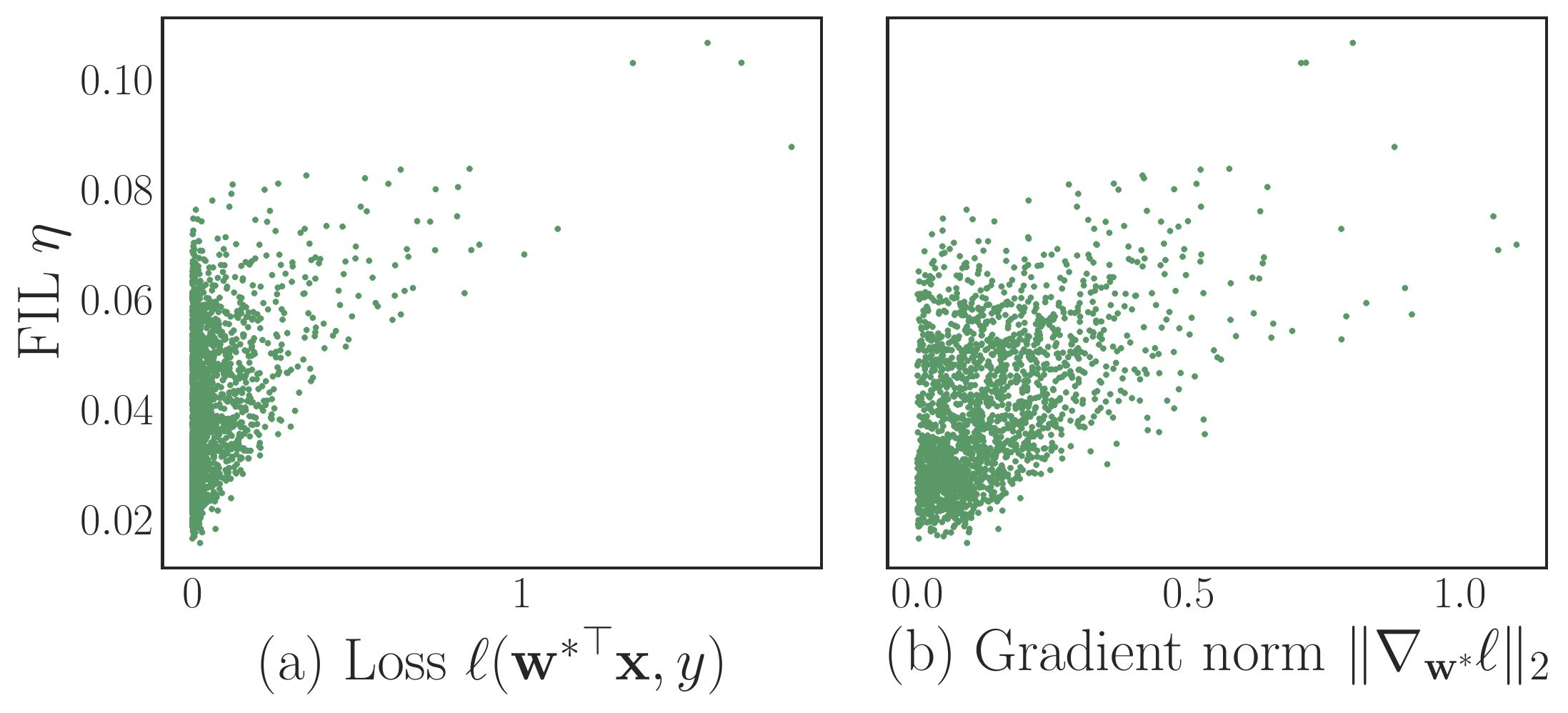}
  \caption{Scatter plots of the FIL $\eta$ with (a) the loss, and (b) the norm
  of the gradient on MNIST with linear regression and 2,000 randomly selected
  examples.}
  \label{fig:correlation_alternatives}
\end{figure}

We compare $\eta$ to alternative heuristics that could correlate
to the information a model contains about an example. Specifically, we measure
the per-example loss $\ell(\vw^{*\top} \vx_i, y_i)$, and the norm of the
gradient $\|\nabla_{\vw^*} \ell(\vw^{*\top} \vx_i, y_i)\|_2$.
Figure~\ref{fig:correlation_alternatives} shows scatter plots of $\eta$ against
these alternatives. The FIL $\eta$ correlates with both metrics; however, high
values of $\eta$ exist throughout the range of both alternatives. On the other
hand, a low $\eta$ invariably implies a low loss. This shows that while
loss-based heuristics may enjoy high precision in assessing privacy
vulnerability, such heuristics should be used with caution, since they could
suffer from low recall.

\begin{figure}
  \begin{subfigure}{0.49\linewidth}
  \centering
  \includegraphics[width=1.0\textwidth]{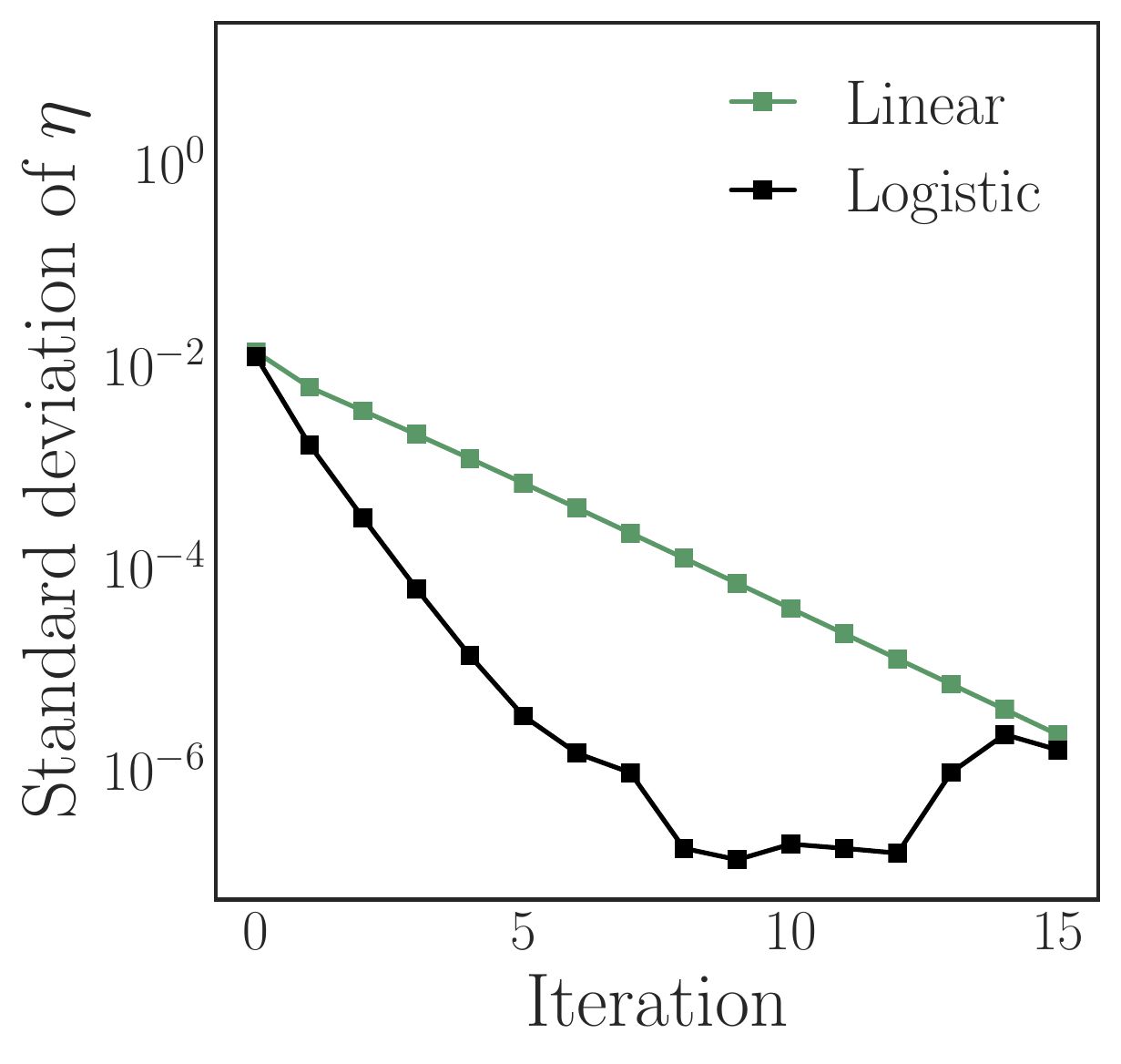}
    \caption{MNIST}
  \end{subfigure}
  \begin{subfigure}{0.49\linewidth}
  \centering
  \includegraphics[width=1.0\textwidth]{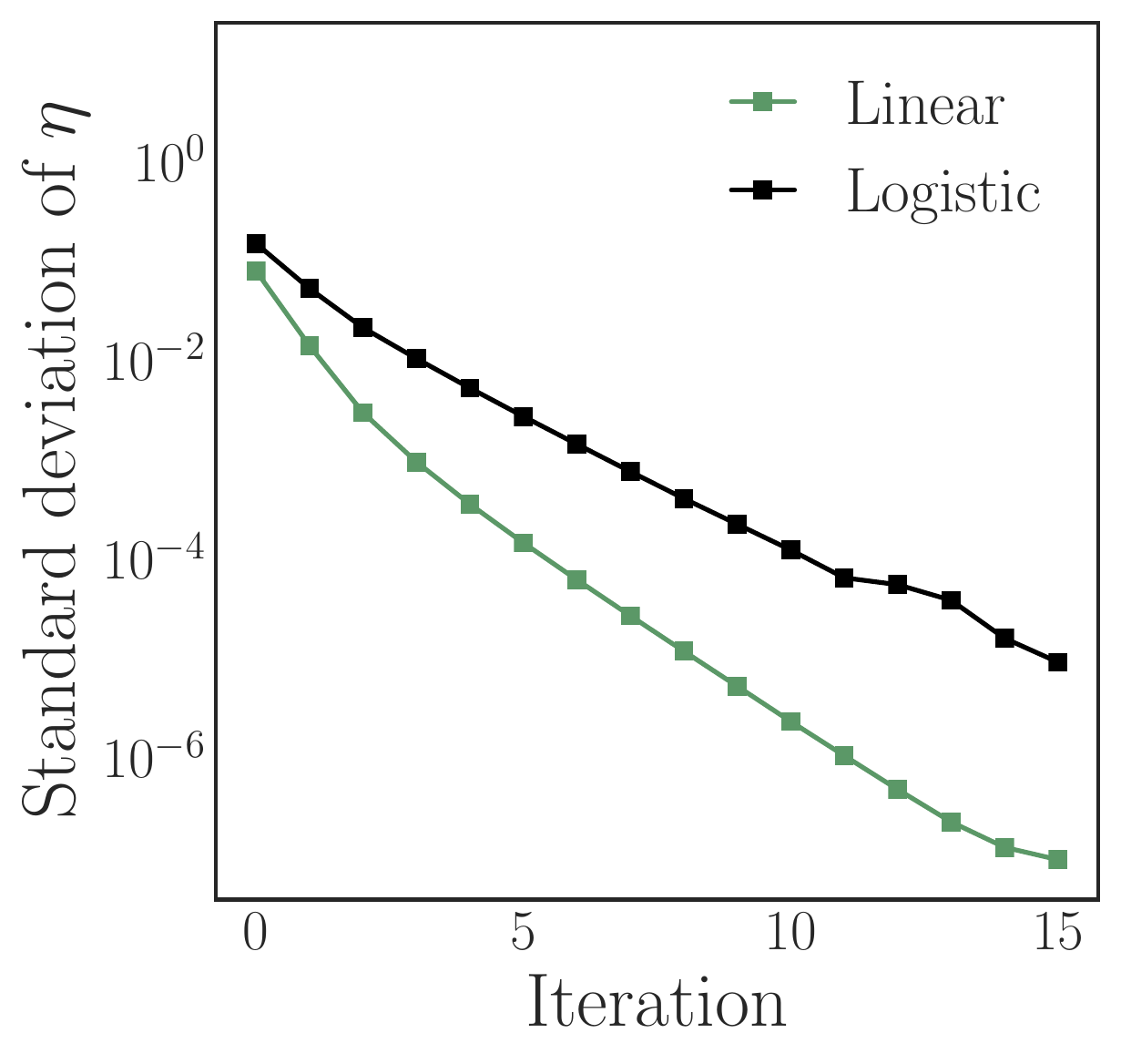}
    \caption{CIFAR-10}
  \end{subfigure}
  \caption{The standard deviation of the example-level $\eta$ over
  iterations of the IRFIL algorithm.}
  \label{fig:irfil_eta}
\end{figure}

\begin{figure*}[t]
  \centering
  \begin{subfigure}{0.24\textwidth}
  \centering
  \includegraphics[width=1.0\textwidth]{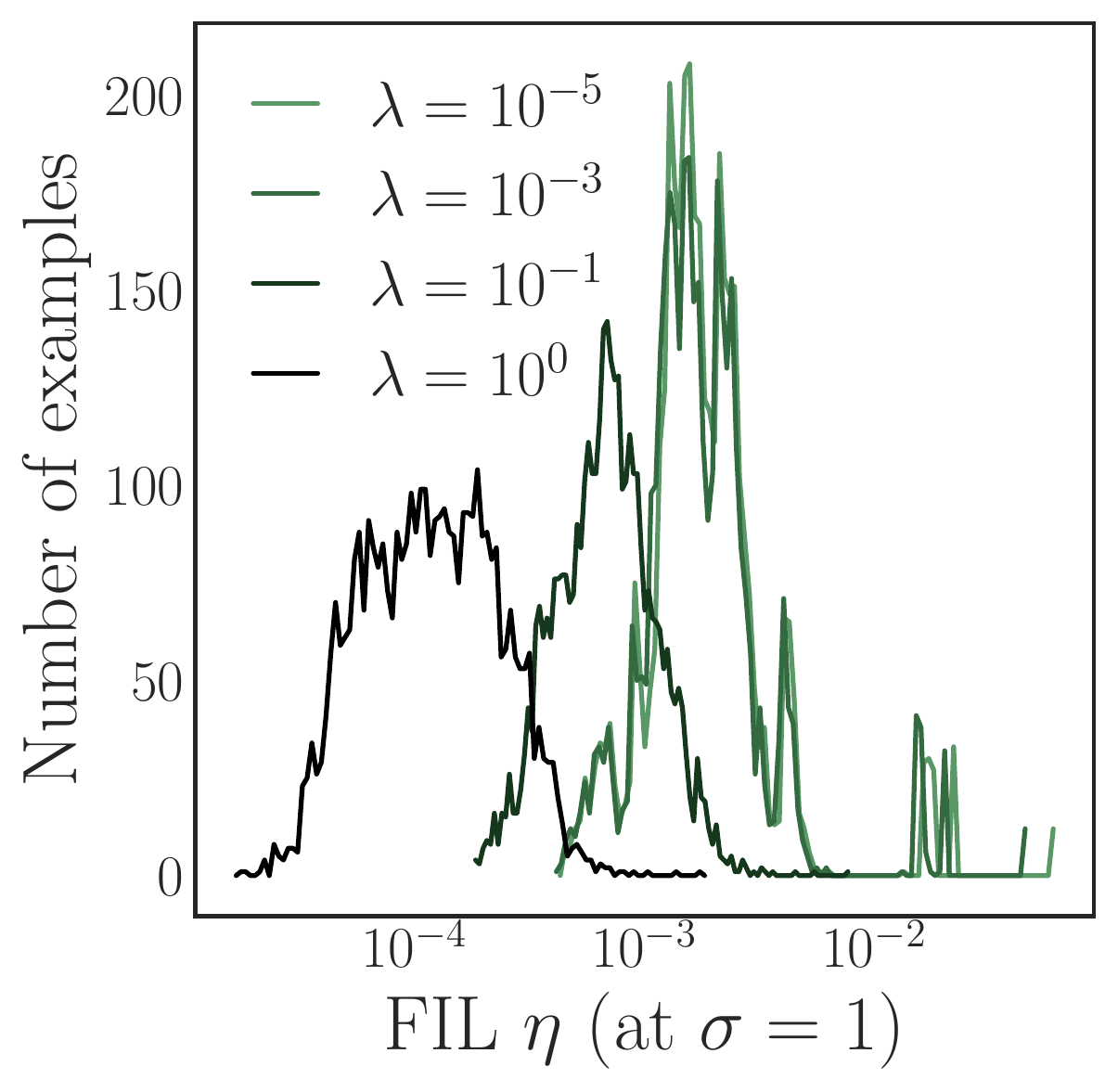}
  \caption{Distribution of $\eta$}
  \end{subfigure}
  \begin{subfigure}{0.24\textwidth}
  \centering
  \includegraphics[width=1.0\textwidth]{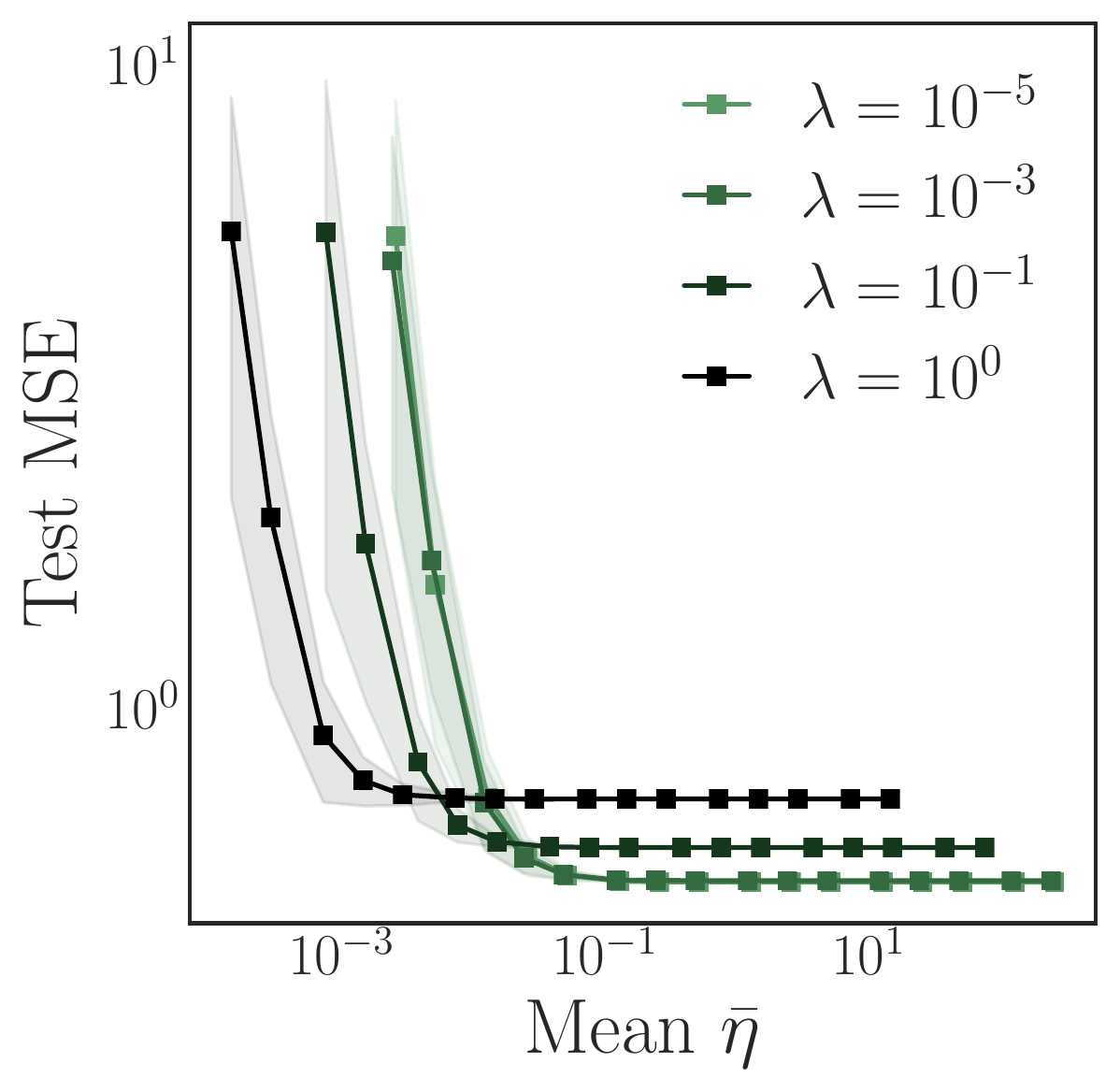}
  \caption{Test MSE}
  \label{fig:iwpc_mse}
  \end{subfigure}
  \begin{subfigure}{0.24\textwidth}
  \centering
  \includegraphics[width=1.0\textwidth]{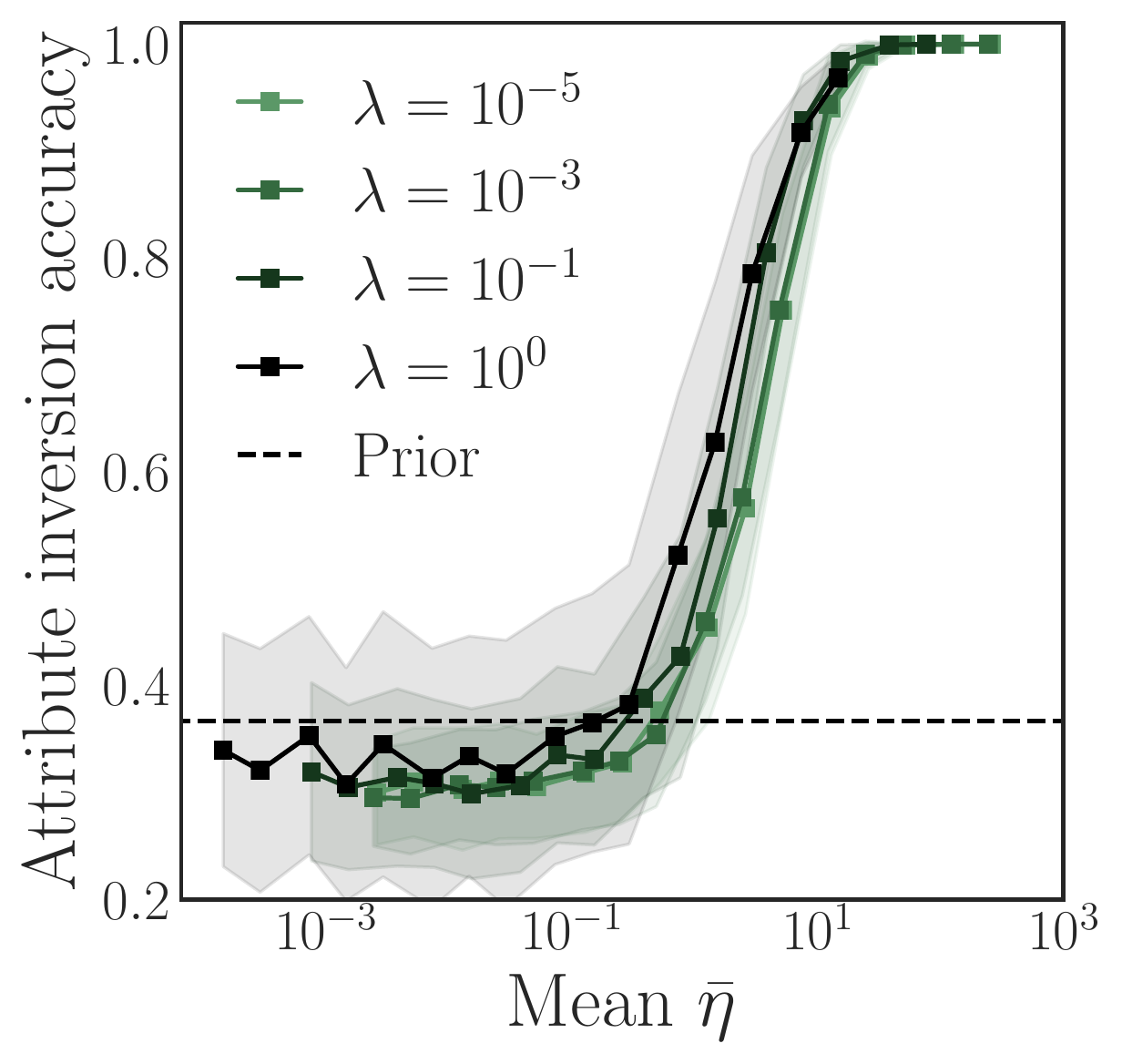}
  \caption{White-box attack}
  \label{fig:iwpc_whitebox}
  \end{subfigure}
  \begin{subfigure}{0.24\textwidth}
  \centering
  \includegraphics[width=1.0\textwidth]{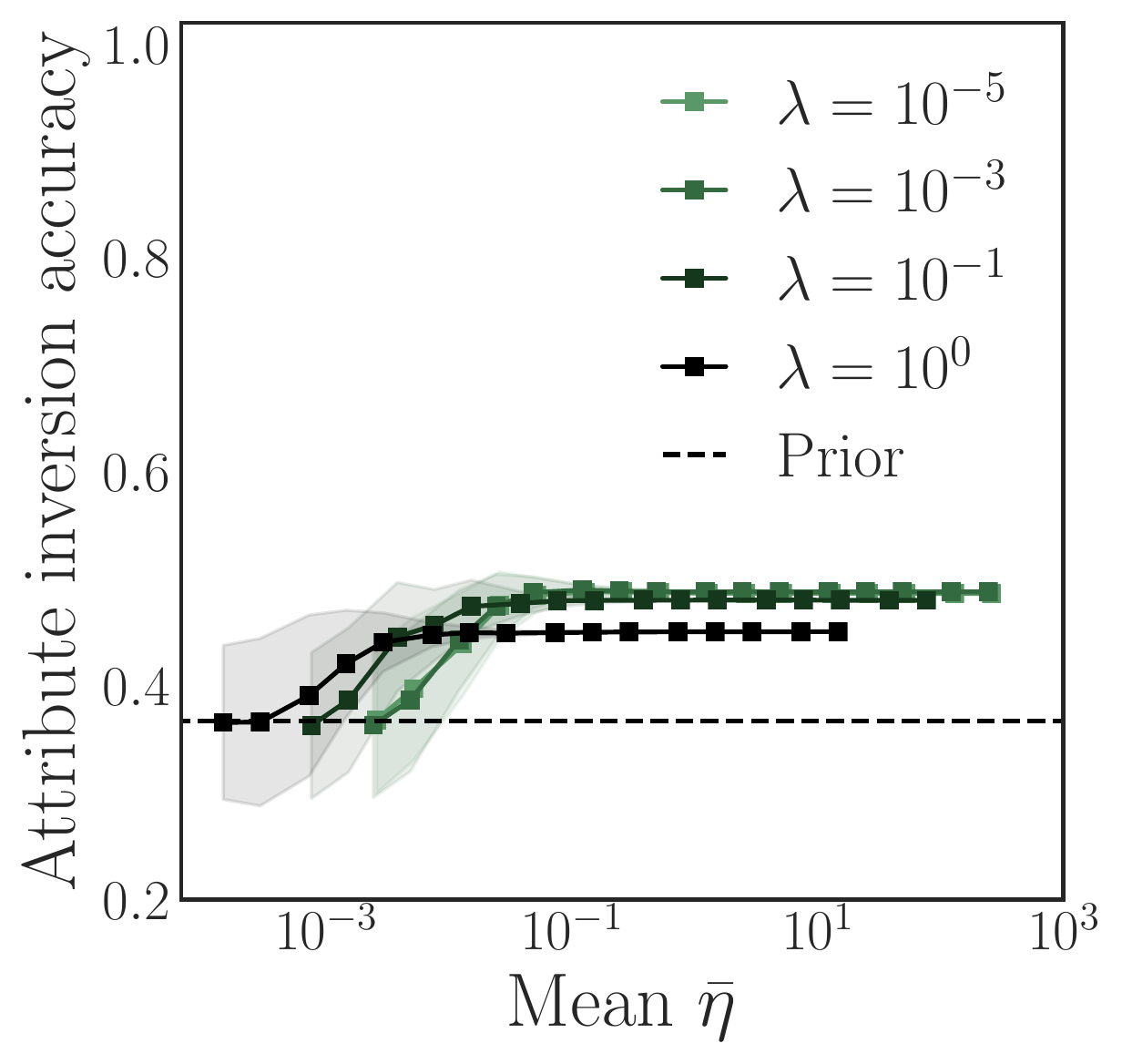}
  \caption{Black-box attack}
  \label{fig:iwpc_blackbox}
  \end{subfigure}
  \caption{The effect of (a) the $L_2$ regularization parameter $\lambda$ on
  $\eta$ for IWPC, and the effect of the mean $\bar{\eta}$ varying $\lambda$ on
  (b) test MSE, (c) white-box accuracy, and (d) black-box accuracy. Standard
  deviations are computed over $100$ trials.}
  \label{fig:iwpc_attr_inversion}
\end{figure*}

\begin{figure*}[t]
  \centering
  \begin{subfigure}{0.243\textwidth}
  \centering
  \includegraphics[width=1.0\textwidth]{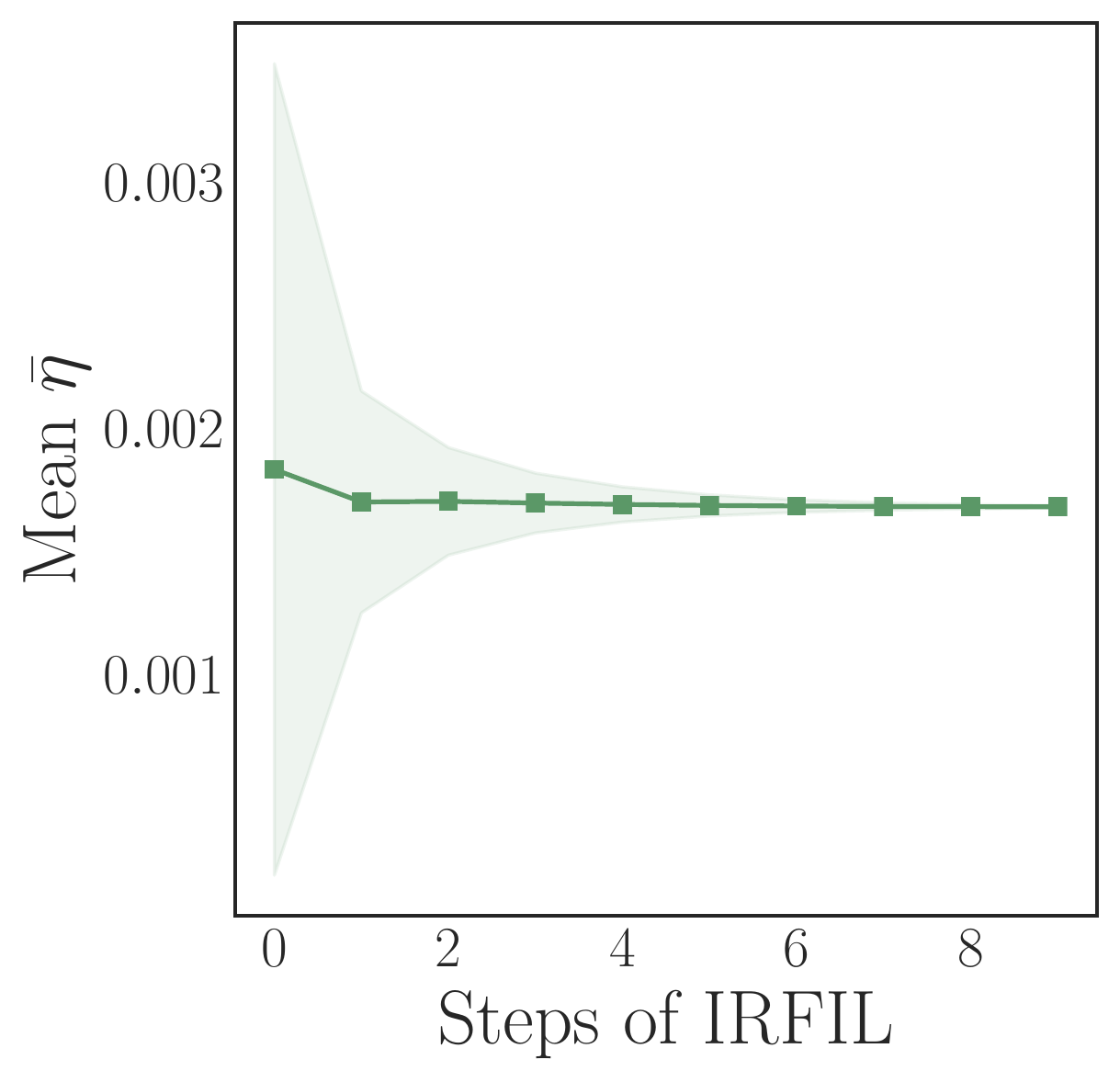}
    \caption{IWPC; $\eta$ mean ($\pm$ SD)}
  \end{subfigure}
  \begin{subfigure}{0.24\textwidth}
  \centering
  \includegraphics[width=1.0\textwidth]{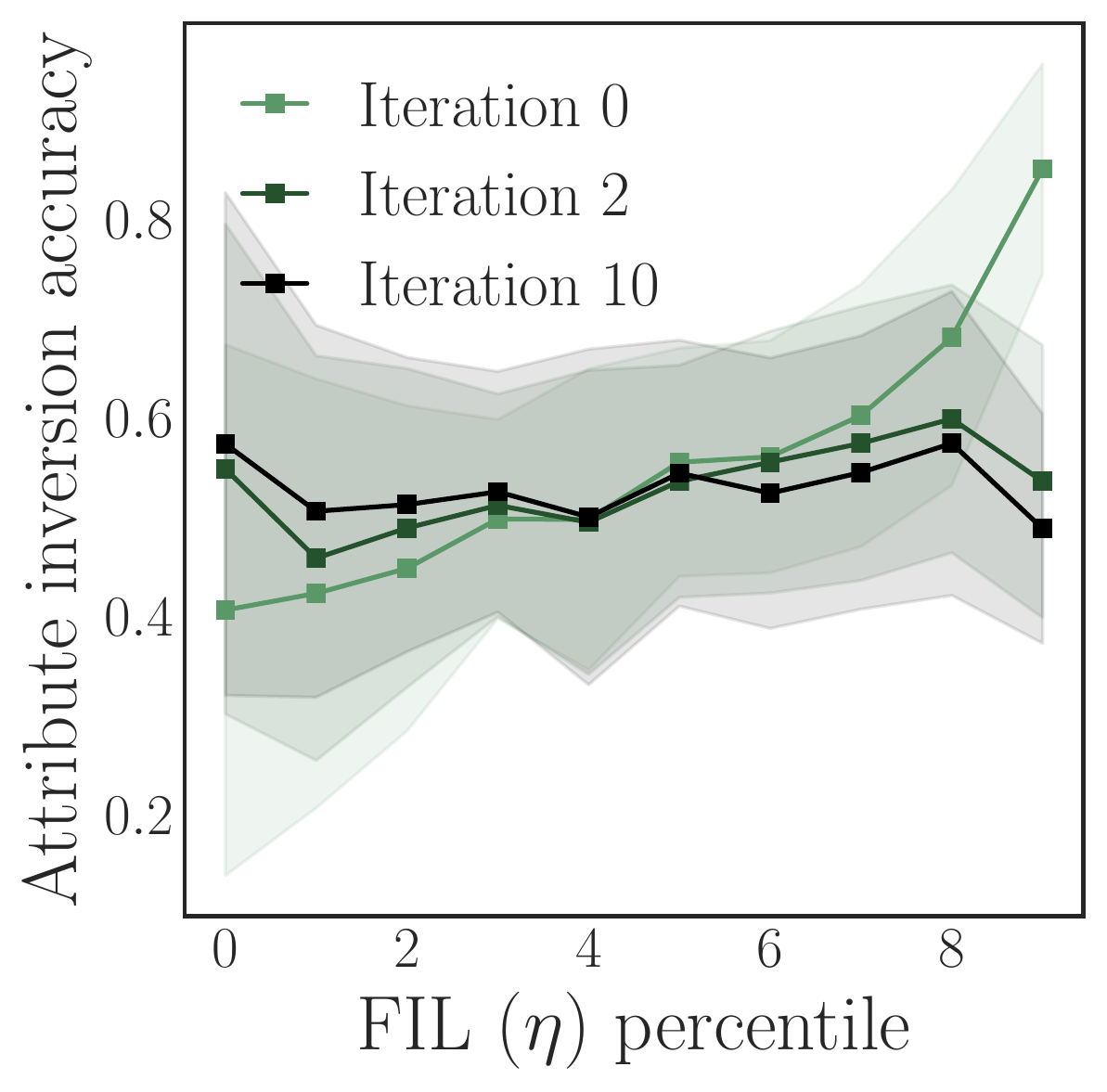}
  \caption{IWPC; white-box accuracy}
  \end{subfigure}
  \begin{subfigure}{0.243\textwidth}
  \centering
  \includegraphics[width=1.0\textwidth]{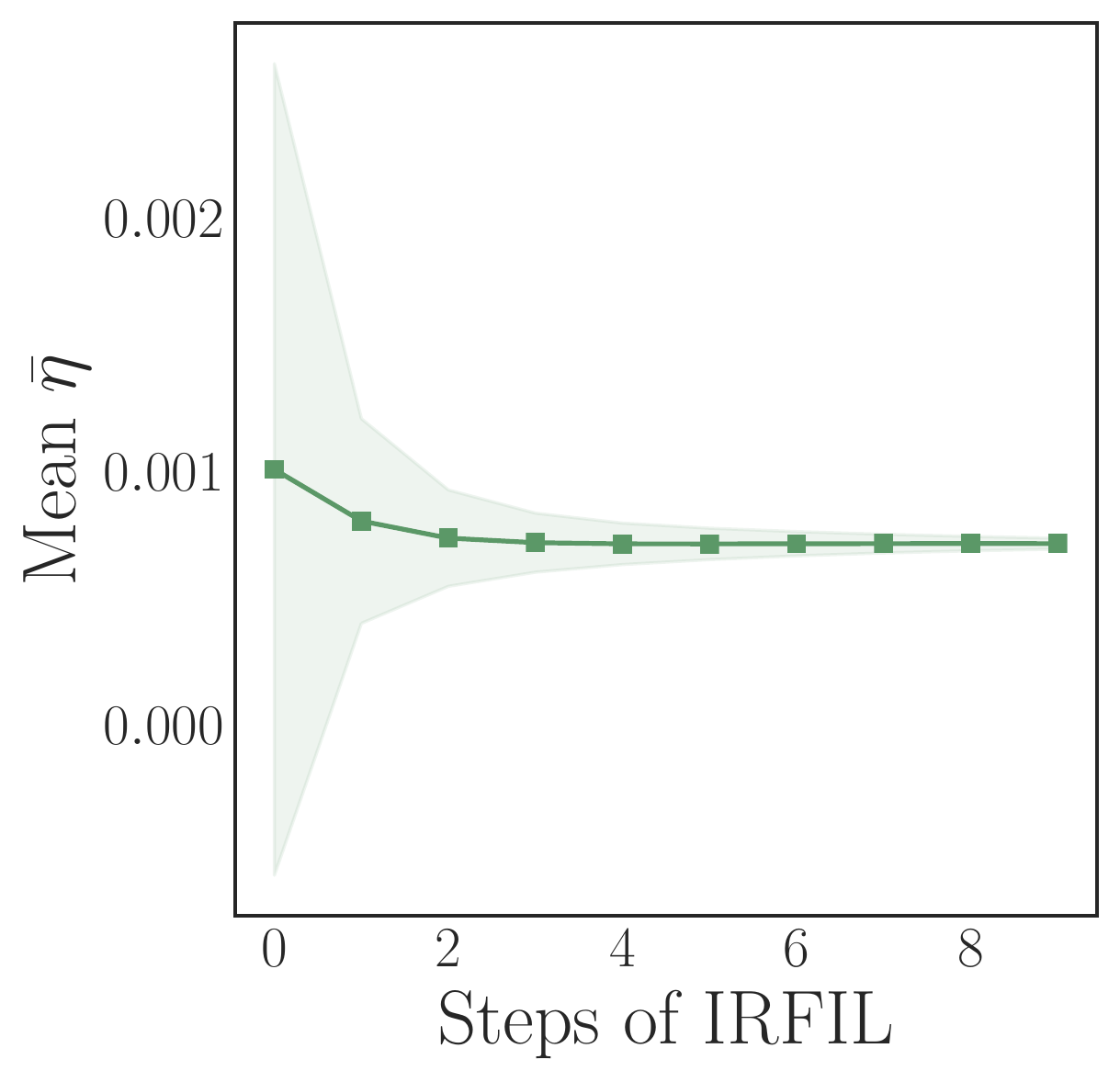}
    \caption{UCI; mean $\bar{\eta}$ ($\pm$ SD)}
  \end{subfigure}
  \begin{subfigure}{0.24\textwidth}
  \centering
  \includegraphics[width=1.0\textwidth]{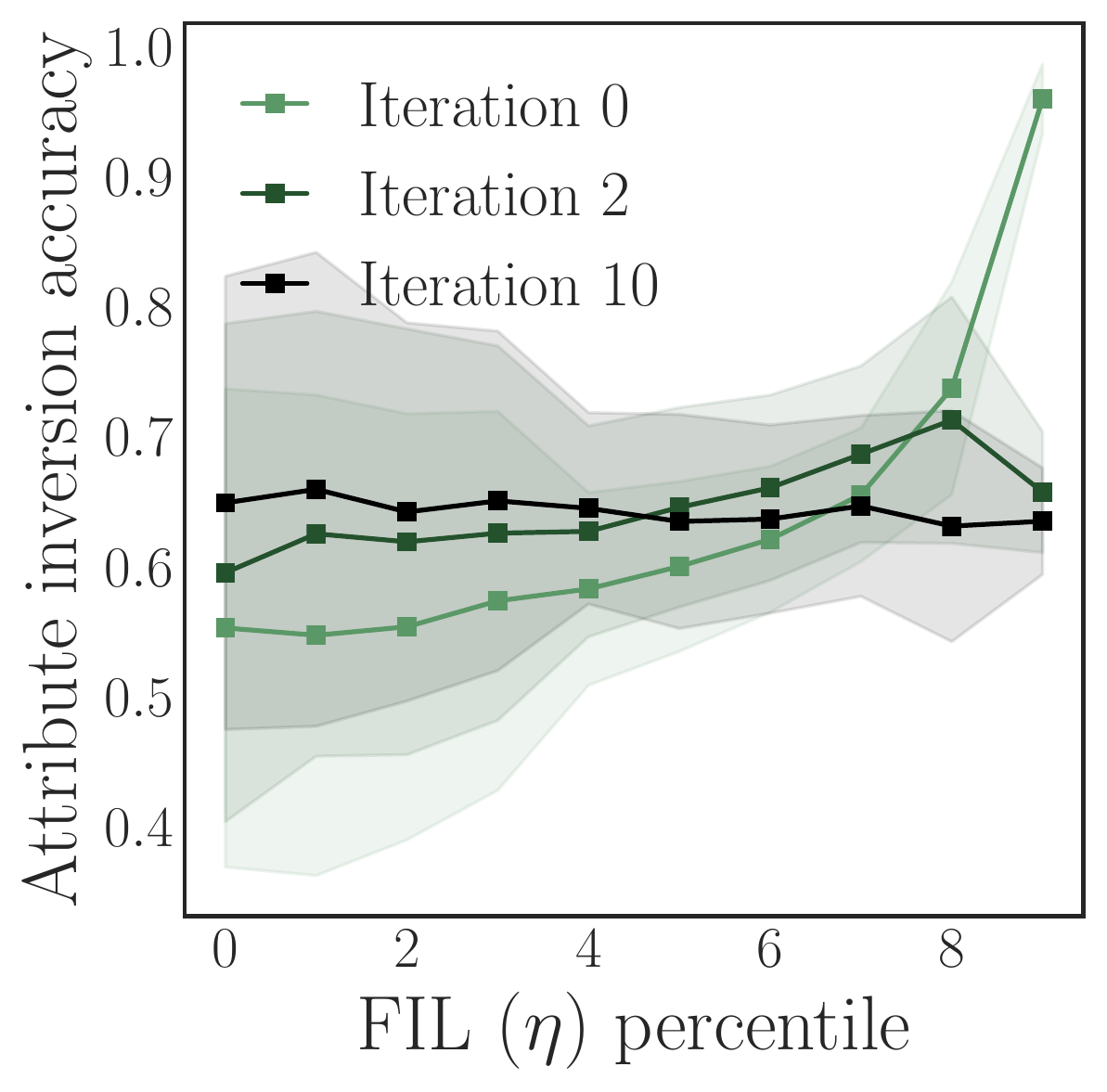}
  \caption{UCI; white-box accuracy}
  \end{subfigure}
  \caption{We show how the per-example attribute-level $\eta$ converge over
  iterations of IRFIL for (a) IWPC and (c) UCI Adult. The mean accuracy of the
  white-box attack over increasing $\eta$-deciles also converges to a similar
  value over iterations of IRFIL for (b) IWPC and (d) UCI Adult. Standard
  deviations are computed over $100$ trials.}
  \label{fig:irfil_attribute_inversion}
\end{figure*}

\subsection{Reweighted FIL}

We empirically evaluate the IRFIL algorithm in \figref{fig:irfil_eta}, which
plots the standard deviation of the per-example $\eta$ against the number of
re-weighting iterations. The per-example $\eta$ converge to the same value
after only a few iterations for both linear and logistic regression on MNIST
and CIFAR-10. Table~\ref{tab:eta_label_mean} shows the mean and standard
deviation of $\eta$, as well as the test accuracy for models trained with and
without IRFIL. Neither the average FIL $\bar{\eta}$ nor the test accuracy are
especially sensitive to the IRFIL algorithm. However, without IRFIL the
standard deviation in $\eta$ is significantly higher, implying that the initial
information leakage varies substantially across training examples. Overall,
IRFIL achieves fairness in privacy loss with little change in accuracy or
average privacy loss.

\begin{table}
  \setlength{\tabcolsep}{3pt}
  \centering
  \caption{The mean $\bar{\eta}$ ($\pm$ standard deviation) of the example-level
  $\eta$ and test accuracy before and after IRFIL.}
  \centering
  \small
  \begin{tabular}{l c c c c}
  \toprule
  & \multicolumn{2}{c}{MNIST} & \multicolumn{2}{c}{CIFAR-10} \\
  Model    & $\bar{\eta}$ & Accuracy & $\bar{\eta}$ & Accuracy \\
  \midrule
  Linear   & 0.040 $\pm$ 0.014 & 100  & 0.26 $\pm$ 0.08 & 79.8 \\
   +IRFIL  & 0.047 $\pm$ 0.000 & 99.8 & 0.30 $\pm$ 0.00 & 80.7 \\
  Logistic & 0.027 $\pm$ 0.012 & 99.8 & 0.32 $\pm$ 0.15 & 79.9 \\
   +IRFIL  & 0.024 $\pm$ 0.000 & 99.7 & 0.25 $\pm$ 0.00 & 79.1 \\
  \bottomrule
  \end{tabular}
  \label{tab:eta_label_mean}
\end{table}

\subsection{Attribute Inference Attacks}

We investigate how the success of attribute inference attacks varies with
different levels of Fisher information loss. To do so, we use two attribute
inversion attacks: 1) a white-box attack based on the FIL threat model, and 2)
a black-box attack following the method described
in~\citet{fredrikson2014privacy}. The goal of the adversary is to infer the
value of a nominal target attribute $x^\textrm{tgt} \in \vx$ for a given
example.

{\bf White-box attack.} The white-box setting assumes the adversary has access
to the complete training dataset $\gD \setminus x_i^{\textrm{tgt}}$ for all but
the target attribute of the example under attack. The adversary also has access
to the perturbed model parameters $\vw'$. We also assume the adversary has complete
knowledge of the model training details. Only the value of $x_i^{\textrm{tgt}}$
for the example under attack is opaque to the adversary.

The adversary infers the hidden attribute of the $i$-th
example by estimating $\hat{x}_i^\textrm{tgt}$ to minimize the distance of the
derived model from the given model $\vw'$:
\begin{equation}
  \hat{x}_i^\textrm{tgt} = \argmin_{x^{\textrm{tgt}}}
    \left\|\vw' - f(\gD \setminus x_i^{\textrm{tgt}} \cup x^{\textrm{tgt}}) \right\|_2,
\end{equation}
where $f(\cdot)$ yields the minimizer (\eqref{eq:erm}).

{\bf Black-box attack.} In the black-box setting the adversary has access to
the target example except the value of the target attribute $\vx_i \setminus
x_i^{\textrm{tgt}}$, the label $y_i$, the prior distribution
$p(x^{\textrm{tgt}})$, model predictions $\phi(\vw', \vx)$ via black-box
queries, and model performance statistics $\pi(\phi(\vw', \vx), y)$. For a
given model the attack infers the target attribute value by maximizing:
\begin{equation}
  \hat{x}_i^{\textrm{tgt}} = \argmax_{x^{\textrm{tgt}}} p(x^{\textrm{tgt}})
    \pi(\phi(\vw', \vx_i \setminus x_i^\textrm{tgt} \cup x^\textrm{tgt}), y_i).
\end{equation}
We compare both attacks to a simple baseline adversary which infers the target
attribute $\hat{x}_i^{\textrm{tgt}} = \argmax_{x^{\textrm{tgt}}}
p(x^{\textrm{tgt}})$.

For patients in IWPC, we infer the allele of the VKORC1 gene, which has three
possible values. We use linear regression with $\lambda = 10^{-2}$.  For the
black-box adversary, the performance metric $\pi(\phi(\vw', \vx), y) =
p(\vw'^\top \vx; y, s^2)$ is a Gaussian distribution with mean $y$
and variance given by the standard error $s^2$ on the training set.

{\bf Result.} Figure~\ref{fig:iwpc_attr_inversion} shows the effect of $L_2$
regularization and the mean $\bar{\eta}$ on the accuracy of the white-box and
black-box attacks. For larger $\lambda$, the MSE degrades at smaller
$\bar{\eta}$ (\ref{fig:iwpc_mse}), suggesting more $L_2$ regularization
requires less noise (smaller $\sigma$) to achieve the same $\bar{\eta}$. The
white-box (\ref{fig:iwpc_whitebox}) and black-box (\ref{fig:iwpc_blackbox})
accuracies degrade as $\eta$ decreases. However, accross $\lambda$ the
white-box accuracy follows $\bar{\eta}$ more closely than the black-box
accuracy (\ref{fig:iwpc_blackbox}), which tends to track with MSE. By choosing
$\sigma$ such that $\bar{\eta} \approx 10^{-3}$, we can protect against both
adversaries with little degradation in MSE.

We show the effect of IRFIL on the white-box attack using both the IWPC and UCI
Adult datasets. For UCI Adult, we infer the binary marital status attribute.
For both datasets, we compute $\eta$ specific to the attribute under attack for
each example and use $\sigma=10^{-3}$.
Figure~\ref{fig:irfil_attribute_inversion} shows that the $\eta$ for each
example rapidly converge to the same value over iterations of IRFIL. We sort
the examples into deciles based on their initial $\eta$ values. We then compute
the average attribute inversion accuracy for each decile at iterations of
IRFIL.  For both datasets, the highest $\eta$ decile is initially much more
susceptible to attribute inversion than the lowest $\eta$ decile. However,
after only two iterations of IRFIL, the inversion accuracies flatten
substantially across $\eta$ deciles, and after ten iterations they are nearly
constant. Hence, IRFIL can equalize vulnerability to privacy attacks across
individuals.

\section{Discussion and Future Work}
\label{sec:discussion}

We demonstrated that Fisher information loss can be used to assess the
information leaked by a model about its training data. A primary benefit of FIL
over \emph{a priori} guarantees like differential privacy is the ability to
measure information leakage at various granularities with respect to the data
at hand. This also allows FIL to be used to construct models with
equi-distributed leakage, which can be done with iteratively reweighted FIL.
Furthermore, FIL explicitly measures the inferential power of an adversary, and
we validated that it correctly captures vulnerability to privacy attacks.
As a result, FIL can be used by practitioners to tailor the resulting privacy
to the desired granularity and to the adversary's knowledge and capabilities.

We motivated the use of FIL via the Cram\'{e}r-Rao bound and delineated the
corresponding threat model. However, the assumption that the adversary is limited to
unbiased estimators may not hold in the presence of auxiliary information. The
implications of this should be further investigated. Furthermore, unlike
differential privacy, FIL does not implicitly degrade with correlated data.
This property of FIL should also be further studied.

The IRFIL algorithm closely resembles iteratively reweighted least squares
(IRLS), which has been widely studied for $\ell_p$-norm
regression~\citep{green1984iteratively, burrus1994iterative} and sparse
recovery~\citep{daubechies2010iteratively}. While IRLS often converges rapidly
in practice, the theoretical convergence rates are difficult to derive and do
not reflect the empirical results~\citep{ene2019improved}. We also observed
rapid and robust convergence with the IRFIL algorithm without any
hyper-parameter tuning. Future work may help understand the convergence of IRFIL
from a theoretical standpoint.

Finally, we considered FIL in the common setting of output-perturbed generalized
linear models with Gaussian noise. However, many possible extensions exist in the
randomization used including alternative noise distributions such as the
Laplace distribution~\citep{dwork2006calibrating}, objective
perturbation~\citep{chaudhuri2011}, or quantifying leakage via predictions
directly using, for example, the exponential mechanism~\citep{mcsherry2007}.
Furthermore, extending FIL to the setting of non-linear and non-convex models
will facilitate its utility and broader adoption.

\begin{acknowledgements}
  Thanks to Mark Tygert for essential contributions to the ideas in this work
  and for providing copious feedback.
\end{acknowledgements}

\bibliography{references.bib}

\appendix
\section{\uppercase{Fisher Information of the Gaussian Mechanism}}
\label{apx:fim_gaussian_derivation}

We provide a simple derivation of the FIM of the Gaussian mechanism applied to
the empirical risk minimizer, $\vw^*$. The conditional probability density of
the output perturbed parameters is given by:
\begin{equation}
  \label{eq:output_perturb_dist}
  p(\vw' \mid \gD) = \int_{\vw^*} p(\vw' \mid \vw^*, \gD) p(\vw^* \mid \gD) d\vw^* = p(\vw' \mid \vw^*)
\end{equation}
where in the last step we use the fact that $\vw^*$ is sufficient for $\vw'$.
We also assume $f(\gD)$ is deterministic, and hence $p(\vw^* \mid \gD)$ is a
(shifted) delta function nonzero at the optimal parameters, $\vw^*$.

Using~\eqref{eq:output_perturb_dist}, the gradient of $\log p(\vw' \mid \gD)$
with respect to $\gD$ is given by:
\begin{equation}
  \nabla_{\gD} \log p(\vw' \mid \gD) = \mJ_f^\top \nabla_{\vw^*} \log p(\vw' \mid \vw^*)
\end{equation}
where $\mJ_f$ is the Jacobian of $f(\gD)$ with respect to $\gD$. The Hessian is:
\begin{equation}
  \label{eq:output_perturb_fisher_second_order}
  \begin{split}
  &\nabla^2_\gD \log p(\vw' \mid \gD) = \\
    &\mJ_f^\top \nabla^2_{\vw^*} \log p(\vw' \mid \vw^*) \mJ_f + \tH \nabla_{\vw^*} \log p(\vw' \mid \vw^*)
  \end{split}
\end{equation}
where $\tH$ is the three-dimensional tensor of second-order derivatives (in a
slight abuse of notation $\tH_{ijk} = \frac{\partial^2 f_k}{\partial \gD_i
\gD_j}$). Using the second-order expression for the FIM requires evaluating
the expectation over $\vw'$ of~\eqref{eq:output_perturb_fisher_second_order}.

When using zero-mean isotropic Gaussian noise for the perturbation, $\gN(0,
\sigma^2 \mI)$, the expectation over $\vw'$ of
\eqref{eq:output_perturb_fisher_second_order} simplifies. The gradient of $\log
p(\vw' \mid \vw^*)$ is:
\begin{equation}
  \nabla_{\vw^*} \log p(\vw' \mid \vw^*) = \frac{\vw' - \vw^*}{\sigma^2},
\end{equation}
and hence the Hessian is:
\begin{equation}
  \nabla^2_{\vw^*} \log p(\vw' \mid \vw^*) = -\frac{1}{\sigma^2}\mI.
\end{equation}
Evaluating the expectation of \eqref{eq:output_perturb_fisher_second_order}
using the above expressions yields:
\begin{align*}
  &\E\left[\mJ_f^\top \nabla^2_{\vw^*} \log p(\vw' \mid \vw^*) \mJ_f + \tH \nabla_{\vw^*} \log p(\vw \mid \vw^*)\right] = \\
  &\mJ_f^\top \E \left[ \nabla^2_{\vw^*} \log p(\vw' \mid \vw^*) \right] \mJ_f + \tH \E \left[ \nabla_{\vw^*} \log p(\vw' \mid \vw^*)\right] = \\
  &-\frac{1}{\sigma^2} \mJ_f^\top \mJ_f,
\end{align*}
where the second term vanishes since $\E[\vw'] = \vw^*$. Hence the FIM is given
by:
\begin{equation}
\gI_{\vw'}(\gD) = -\E\left[ \nabla^2_\gD \log p(\vw' \mid \gD) \right] = \frac{1}{\sigma^2}\mJ_f^\top \mJ_f.
\end{equation}

\section{\uppercase{Jacobian of the Minimizer}}
\label{apx:jacobian_general}

Let $\ell(\vw^\top \vx, y)$ be a convex, twice-differentiable loss function. Let
$f_i(\vx, y)$ denote the minimizer of the regularized empirical risk as a
function of $(\vx, y)$ at the $i$-th example:
\begin{equation}
  \label{eq:erm_example_2}
f_i(\vx, y) = \argmin_{\vw} \sum_{j \neq i} \ell(\vw^\top \vx_j, y_j) + \ell(\vw^\top \vx, y) + \frac{n \lambda}{2} \| \vw \|_2^2.
\end{equation}
We aim to derive an expression for $\mJ_{f_i}\big\rvert_{x_i, y_i}$, the
Jacobian of $f_i(\vx, y)$ with respect to $(\vx, y)$ evaluated at $(\vx_i,
y_i)$.  Taking the gradient of \eqref{eq:erm_example_2} with respect to $\vw$
and setting it to $0$ gives an implicit function for $\vw^* = f_i(\vx, y)$:
\begin{equation}
  0 = \sum_{j \neq i} \nabla_\vw \ell(\vw^{*\top} \vx_j, y_j) + \nabla_\vw \ell(\vw^{*\top} \vx, y) + n \lambda \vw^*.
\label{eq:implicit_func}
\end{equation}
Implicit differentiation of \eqref{eq:implicit_func} with respect to $(\vx, y)$
gives:
\begin{equation}
  0 = \sum_{j \neq i} \nabla_\vw^2 \ell(\vw^{*\top} \vx_j, y_j) \mJ_{f_i} + \nabla_{\vx, y} \nabla_\vw \ell(\vw^{*\top}\vx, y) + n \lambda \mJ_{f_i}.
\label{eq:implicit_diff}
\end{equation}
The second term can be computed using the product rule:
\begin{align}
  \nabla_{\vx, y}& \nabla_\vw \ell(\vw^{*\top} \vx, y) = \nonumber \\
  &\nabla_\vw^2 \ell(\vw^{*\top}\vx, y) \mJ_{f_i} + \nabla_{\vx, y} \nabla_\vw \ell(\vw^\top \vx, y) \bigg\rvert_{\vw = \vw^*}.
  \label{eq:product_rule}
\end{align}
Evaluating \eqref{eq:product_rule} at $(\vx_i, y_i)$ and substituting into
equation \ref{eq:implicit_diff} yields:
\begin{align}
  &0 = \nonumber \\
  &\left[\sum_{j = 1}^n \nabla_\vw^2 \ell(\vw^{*\top}\vx_j, y_j) \mJ_{f_i} + \nabla_{\vx, y} \nabla_\vw \ell(\vw^\top\vx, y) + n \lambda \mJ_{f_i}\right]_{\vw^*, \vx_i, y_i} \nonumber \\
  &= \bigg[ \mH_{\vw^*} \mJ_{f_i} + \nabla_{\vx, y} \nabla_\vw \ell(\vw^\top \vx, y) \bigg]_{\vw^*, \vx_i, y_i},
\end{align}
where the Hessian $\mH_{\vw^*} = \sum_{j = 1}^n \nabla_\vw^2 \ell(\vw^{*\top}
\vx_j, y_j) + n\lambda \mI$. Solving for $\mJ_{f_i}$ yields:
\begin{equation}
  \mJ_{f_i}\bigg\rvert_{\vx_i, y_i} = -\mH_{\vw^*}^{-1} \nabla_{\vx, y} \nabla_\vw \ell(\vw^{*\top} \vx_i, y_i).
\end{equation}

\end{document}